\documentclass[letterpaper]{article} 
\usepackage{aaai25}
\usepackage{times}  
\usepackage{helvet}  
\usepackage{courier}  
\usepackage[hyphens]{url}  
\usepackage{graphicx} 
\usepackage{amsmath}
\usepackage{amssymb}
\usepackage{booktabs}
\usepackage{array}
\usepackage{graphicx}
\usepackage{subfigure}
\usepackage{lipsum}
\usepackage{float} 
\usepackage{placeins}
\usepackage{natbib}  
\usepackage{caption} 
\usepackage{algorithm}
\usepackage{algorithmic}

\usepackage{newfloat}
\usepackage{listings}
\DeclareCaptionStyle{ruled}{labelfont=normalfont,labelsep=colon,strut=off} 
\floatstyle{ruled}
\newfloat{listing}{tb}{lst}{}
\floatname{listing}{Listing}
\title{Revisiting Day-ahead Electricity Price: Simple Model Save Millions}

\author{Linian Wang\textsuperscript{\rm 1},
    Jianghong Liu\textsuperscript{\rm 2},
    Huibin Zhang\textsuperscript{\rm 3},
    Leye Wang\textsuperscript{\rm 1}
}
\affiliations{
    \textsuperscript{\rm 1}Key Lab of High Confidence Software Technologies (Peking University), Ministry of Education \& School of Computer Science, Peking University\\

    \textsuperscript{\rm 2}Wuhan Yuanguang Technology Co., Ltd.\\
    \textsuperscript{\rm 3}Nyocor Co., Ltd.\\
    
    wanglinian@stu.pku.edu.cn, liubird@gmail.com, zhanghuibin2010@gmail.com, leyewang@pku.edu.cn
}

\begin{document}

\maketitle

\begin{abstract}
Accurate day-ahead electricity price forecasting is essential for residential welfare, yet current methods often fall short in forecast accuracy. We observe that commonly used time series models struggle to utilize the prior correlation between price and demand-supply, which, we found, can contribute a lot to a reliable electricity price forecaster. Leveraging this prior, we propose a simple piecewise linear model that significantly enhances forecast accuracy by directly deriving prices from readily forecastable demand-supply values. Experiments in the day-ahead electricity markets of Shanxi province and ISO New England reveal that such forecasts could potentially save residents millions of dollars a year compared to existing methods. Our findings underscore the value of suitably integrating time series modeling with economic prior for enhanced electricity price forecasting accuracy.
\end{abstract}

%

\section{Introduction}

Electricity price forecasting is closely linked to residential welfare and deserves thorough investigation. Modern society relies heavily on a stable supply of electrical energy, but the cost can be a significant financial burden. Ensuring reliable power supply while reducing living costs is vital, especially for low-income populations. In some countries, residents buy electricity directly from the market, while in others, state-owned companies purchase and retail it. Regardless of the form, the electricity price in the power market directly affects residents’ cost. Prices fluctuate throughout the day, leading to varying consumption costs~\cite{bielen2017future}. Advanced energy storage systems allow residents to buy electricity during low-price periods and use stored energy during high-price periods, reducing expenses. This strategy relies on accurate price forecasts, which directly impact financial savings~\cite{zheng2020impact,gil2012forecasting}. Therefore, effective electricity price forecasting is essential for optimizing residents' revenue and enhancing societal welfare.

The day-ahead market is a pivotal part of the electricity market, and this paper focuses on forecasting day-ahead electricity prices. Forecasting electricity prices involves various tasks depending on the specific market: Day-ahead, Intra-day, or Balancing markets~\cite{maciejowska2022forecasting}. Among these, the day-ahead market is the most significant, attracting the majority of traders~\cite{Castelli2020Forecasting}. 

Current methods for forecasting day-ahead electricity prices often underperform. These approaches primarily involve the straightforward adaptation of general Time Series Forecast (TSF) models to the day-ahead electricity market~\cite{lago2021survey}. However, the electricity market is characterized by large price fluctuations, a lack of clear periodicity and temporal non-stationarity. Consequently, existing methods may not be effective~\cite{patel2021energy}.

Current forecasting methods are limited by their inability to capture the strong, variable correlations between prices and supply-demand. The price formation mechanism in the electricity market is clear and well-understood; economics indicates that prices are determined by the balance between supply and demand~\cite{Martin_2013}. Therefore, the correlation between electricity prices and market supply-demand is crucial for accurate forecasting~\cite{ZIEL2016435X-Model}. However, TSF methods struggle with this. Some approaches just ignore the impact of supply-demand variables~\cite{Zeng2022dlinear, nie2022patchtst}, while others trying to use data-driven methods to model these correlations~\cite{liu2023itransformer, zhang2023crossformer}. Nonetheless, we find that in trained models, these correlations do not significantly contribute to  forecasting results. 

\begin{figure}[t]
\centering
\includegraphics[width=\columnwidth]{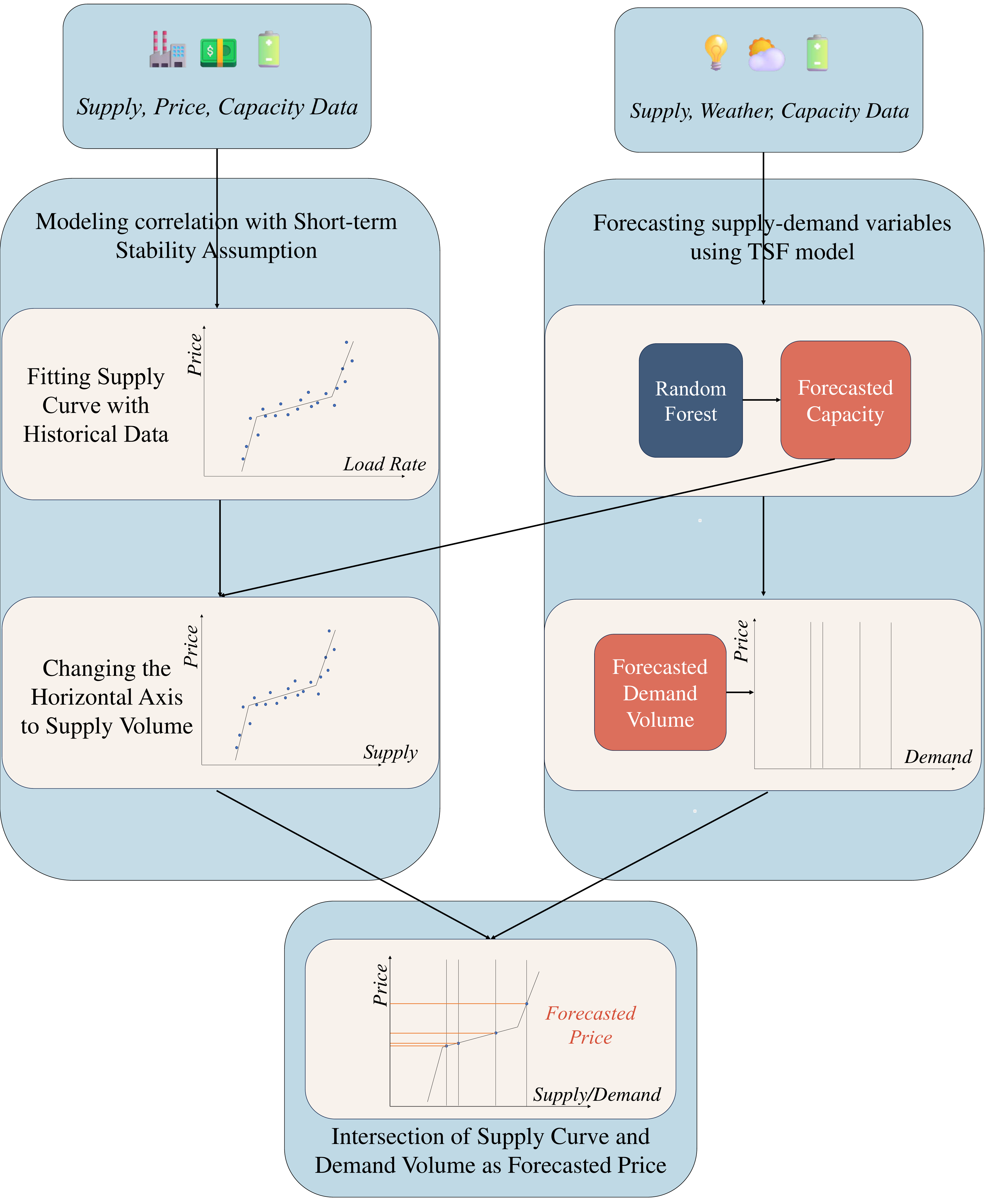} 
\caption{Overview of CoPiLinear.}
\label{fig: overview}
\end{figure}

We propose a simple model to forecast day-ahead electricity prices by modeling the correlation between prices and supply-demand. Instead of directly using TSF models to forecast prices, we utilize supply-related variables forecasted by TSF models and focus on establishing the correlation between these variables and prices based on the price formation mechanism. This approach avoids the weaknesses of current TSF models in capturing variable correlations and provides better interpretability. Specifically, we introduce the assumption of the short-term stability of the supply curve, under which we can forecast the correlation between supply and price using recent historical data. We fit the supply curve using a simple \textbf{Co}rrelation-based \textbf{Pi}ecewise \textbf{Linear} Model (\textbf{CoPiLinear}), then use this and the forecasted demand to forecast the price, as Figure~\ref{fig: overview} shows.

In summary, our main contributions include:

\begin{itemize}
    \item By analyzing the price formation mechanism in the market, we identify the limitations of current models. We introduced a new correlation-based forecasting approach, bridging the gap between the day-ahead electricity price forecasting problem and existing TSF models.
 
    \item To our knowledge, this is one of the first efforts to forecast electricity prices by considering the price formation mechanism in the day-ahead market. We introduce a short-term stability assumption based on market characteristics to simplify supply curve forecasting.
  
    \item Extensive experiments show that our simple method outperforms complex state-of-the-art time (SOTA) models. We demonstrate the revenue of improved electricity price forecasting accuracy for residents, highlighting the significant potential and social impact of the day-ahead electricity price forecasting problem.
 
\end{itemize}

\section{Related Work}

Recent studies on day-ahead electricity price forecasting have employed models directly migrated from existing TSF methods, can be categorized into two types based on the selection of known variables. The first type focuses on the temporal dependency of the target variable, capturing patterns like periodicity and trends. This includes univariate and channel-independent multivariate TSF models. The second type, in addition to considering the temporal dependency of the target variable, also takes into account the correlation between the target variable and auxiliary variables, represented by various channel-dependent multivariate TSF models. 

TSF models emphasize end-to-end forecasting. Advanced second type methods generally capture inter-variable correlations through modules like MLP and attention in parts of the model’s substructure. This approach introduces additional parameters and makes it challenging to intuitively explain the correlations between variables. Moreover, advanced second type methods do not achieve more accurate forecasts with more information as intended. This has been demonstrated by the superior performance of first-type methods like TimesNet~\cite{wu2022timesnet} and PatchTST~\cite{nie2022patchtst} on numerous multivariate datasets.

Modeling the price formation mechanism aligns with economic theory but faces practical challenges due to data unavailability. Accurate projection of supply and demand curves requires data on production and purchase propensities at various price points. However, obtaining comprehensive data is difficult in extensive markets due to the private nature of individual buying and selling decisions, complicating precise measurements of willingness to buy or sell at each price.

\begin{table}[H]
\centering
\resizebox{.95\columnwidth}{!}{
\begin{tabular}{p{3cm}ccc}
\hline
\textbf{Reference} & \textbf{Model} & \textbf{Domain-Specific} & \textbf{Variable Correlation} \\ 
\hline
    \cite{zhao2017improvingsarima} & SARIMA & T & F \\
    \cite{mchugh2019forecastingsarima} & SARIMA & T & F \\
    \cite{xiong2023vmdlstm} & VMD-LSTM & T & F\\
    \cite{Zeng2022dlinear} & DLinear & F & F \\
    \cite{wu2022timesnet} & TimesNet & F & F \\
    \cite{nie2022patchtst} & PatchTST & F & F \\
\hline
    \cite{uniejewski2016automatedlinear} & Linear & T & T\\
    \cite{che2010shortsvm} & SVM & T & T\\
    \cite{wang2017robustsvm} & SVM & T & T\\
    \cite{prahara2022improvedsvm} & SVM & T & T\\
    \cite{manfre2023hybridxgboost} & XGBoost \& LSTM & T & T\\
    \cite{xie2022forecastingxgboost} & XGBoost & T & T\\
    \cite{yamin2004adaptivednn} & DNN & T & T\\
    \cite{darudi2015electricitydnn} & DNN & T & T\\
    \cite{ludwig2015lassorf} & Lasso-RF & T & T \\
    \cite{liu2023koopa} & Koopa & F & T\\
    \cite{chen2023tsmixer} & TSMixer & F & T\\
    \cite{zhou2021informer} & Informer & F & T\\
    \cite{wu2021autoformer} & Autoformer & F & T \\
    \cite{zhou2022fedformer} & FEDformer & F & T \\
    \cite{zhang2023crossformer} & Crossformer & F & T \\
    \cite{liu2023itransformer} & iTransformer & F & T \\
\hline
    \cite{ZIEL2016435X-Model} & Supply-Demand & • & T \\
    \cite{soloviova2020efficient} & Supply-Demand & • & T \\
    \cite{WAN2022105809Nonlinear} & Supply-Demand & • & T \\
    \cite{math10122012Modelling} & Supply-Demand & • & T \\
\hline
\end{tabular}
}
\caption{Summary of related work.}
\label{table:methods}
\end{table}

The summary of related work is listed in Table~\ref{table:methods}. The Model column indicates the primary model used for forecasting in the literature. The Domain-Specific column shows whether the method is tailored for day-ahead electricity price forecasting. A $\bullet$ in the Domain-Specific column denotes methods developed for economic research, not applicable in many markets. The Variable Correlation column reflects if the literature leverages variable correlations for forecasting. 

\section{Preliminary and Problem Formulation}

\begin{figure}[h]
  \centering
  \includegraphics[width=\linewidth]{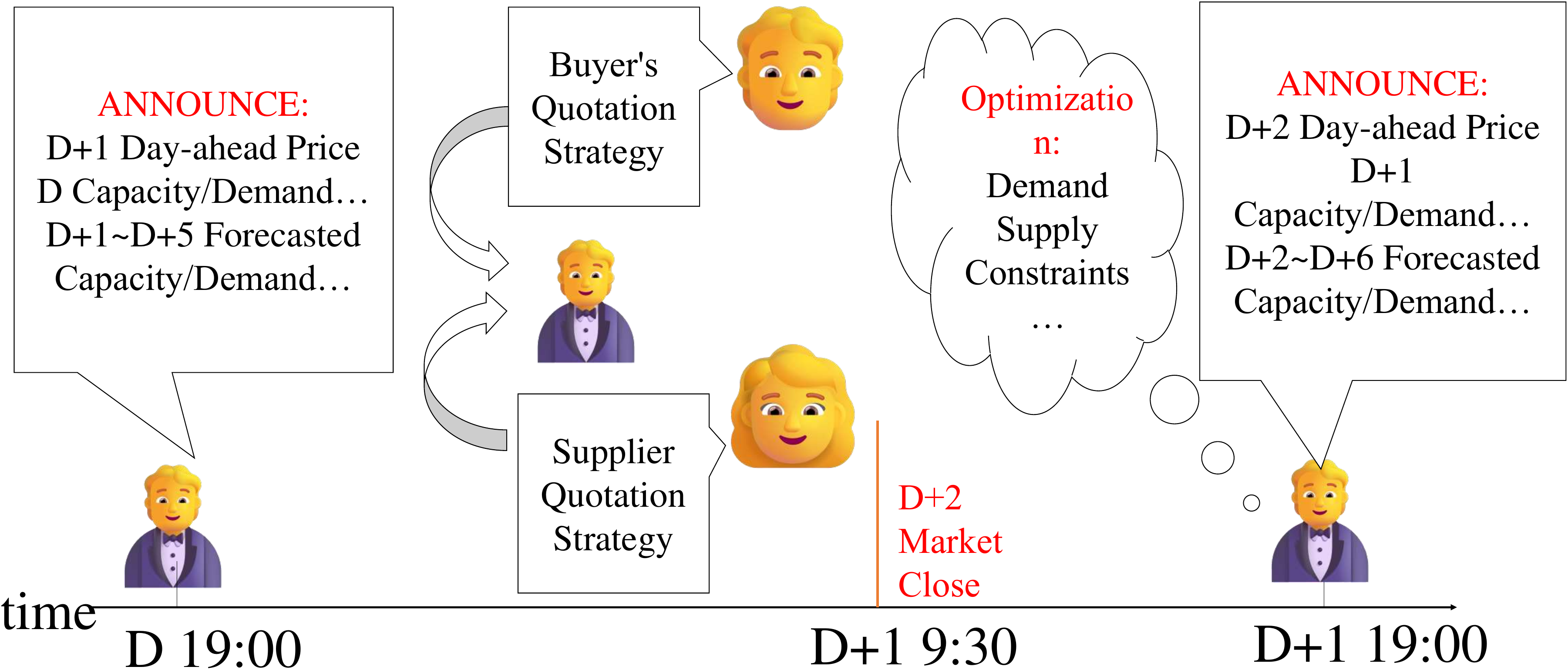}
  \caption{Shanxi market transaction process.}
  \label{fig: market}
\end{figure}

\subsection{Day-ahead Electricity Market Structure}

While regulatory frameworks for day-ahead electricity markets vary by country, their structure generally follows a standard model~\cite{EU2024, Haidar2021}. Prices for each time slot is independently determined through an auction-based system~\cite{Shi2023}, driven by supply and demand dynamics. Take the day-ahead electricity market in Shanxi Province, China, as an example. There are 96 trading time slots in a day, each 15 minutes long. As shown in Figure~\ref{fig: market}, an authoritative third party announces the day-ahead electricity prices for day $D + 1$ and other forecast variables on day D, then market participants submit their bids before the day $D + 2$ market closes. The third party considers anticipated production and costs reported by power producers and accounts for constraints such as grid dispatch limitations, operational characteristics of power plants, and emergency reserves. Through optimization calculations, generation schedules and dispatch plans are coordinated to meet societal electricity demand. Power generation companies then formulate their production plans based on the finalized schedule~\cite{shanxi_power_market}. Through this pricing process of balancing supply and demand, the day-ahead electricity prices for day $D + 2$ in the entire Shanxi market is determined.

\subsection{Limitation of Time Series Forecasting Models}

\begin{figure*}[h]
\centering
\begin{minipage}[t]{0.3\textwidth}
\includegraphics[width=\textwidth]{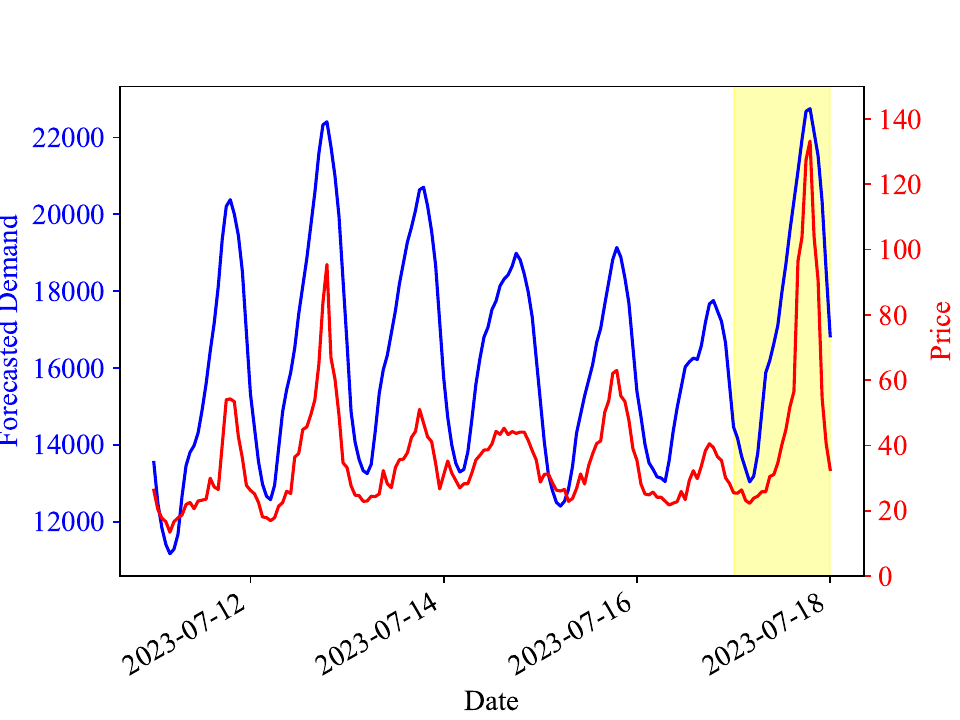}
\caption*{On July 17, 2023 (highlighted in yellow), the forecasted demand peaked for the week, coinciding with an exceptionally high day-ahead electricity price.}
\end{minipage}
\hspace{0.1cm}
\begin{minipage}[t]{0.3\textwidth}
\includegraphics[width=\textwidth]{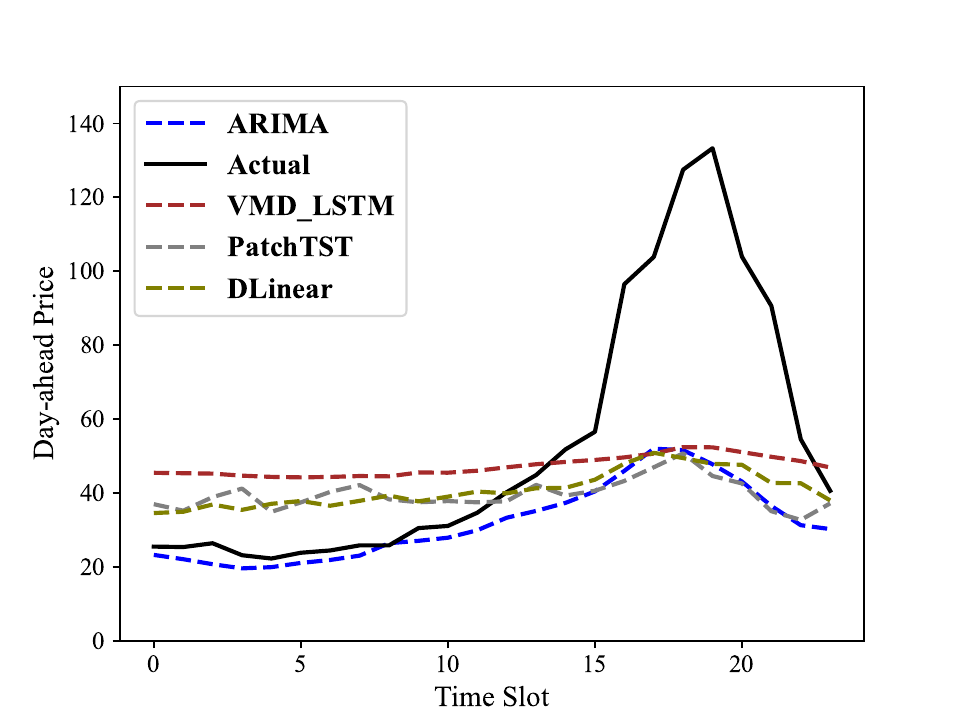}
\caption*{Forecasting models that rely solely on temporal dependence struggle to accurately forecast high day-ahead electricity prices, as seen on July 17, 2023.}
\end{minipage}
\hspace{0.1cm}
\begin{minipage}[t]{0.3\textwidth}
\includegraphics[width=\textwidth]{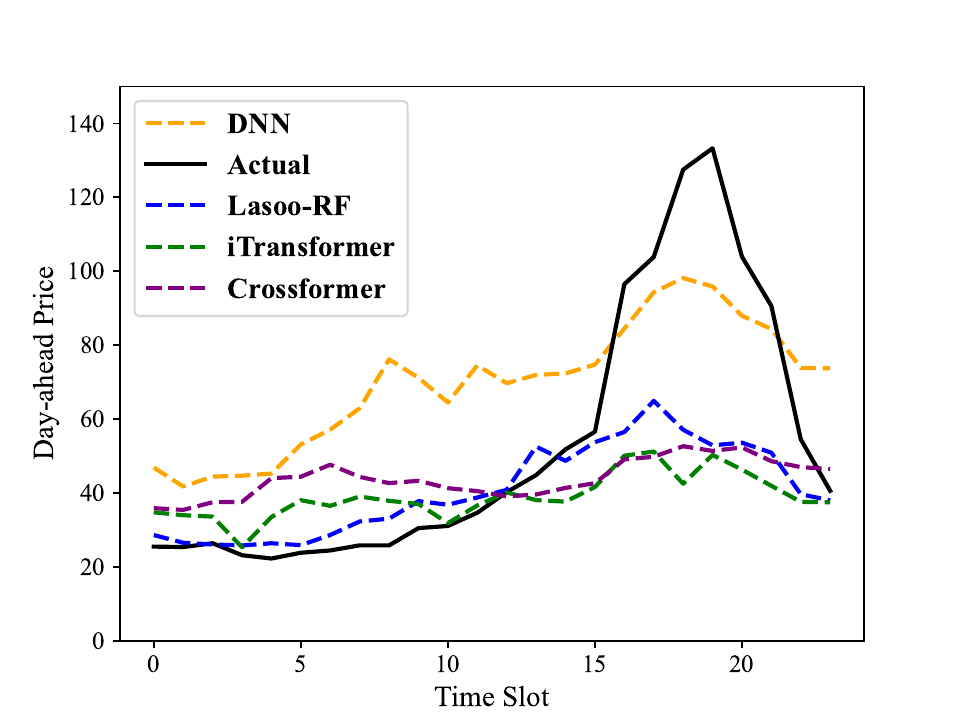}
\caption*{Models that capture variable correlations show relative success in forecasting price increase trends, though the magnitude of these increases is limited.}
\end{minipage}
\caption{Actual and forecasted prices of ISO New England on July 17, 2023.}
\label{fig:shortcoming}
\end{figure*}

Day-ahead electricity prices show significant dispersion in the high-price range. In the ISO New England market, while 75\% of prices are below \$40/MWh, peaks can reach \$300/MWh. This irregularity and lack of clear time series patterns result in poor performance of baseline models, especially time series forecasting models, in forecasting high prices~\cite{rahaman2019spectral,Xu_2020freqpricep}. 

High prices are mainly driven by high demand, which can help forecast these irregularities. However, TSF models struggle to capture the correlation between prices and supply-demand dynamics. According to the price formation mechanism, forecasted demand should significantly impacts forecasted price. Yet, even with substantial changes in demands, existing models show minimal price variation. This inconsistency with real market behavior indicates that these models do not adequately incorporate supply and demand variables, leading to poor forecasting performance. See Appendix A for detailed experiment results.

In Figure~\ref{fig:shortcoming}, the ISO New England Dataset from July 17, 2023, shows forecasted peak demand surpassing the previous week’s peak, indicating high prices. Models based solely on temporal dependence fail to account for forecasted demand, leading to inaccurate high price forecasts. While existing TSF models that capture variable correlations perform better in predicting price trends, they underestimate price increases during significant demand surges, overlooking the price formation mechanism.

\subsection{Problem Formulation}

Our goal is to forecast the day-ahead electricity price for each time slot on day $D+x$ before market closure. The variables at our disposal include day-ahead electricity prices, capacity, weather data and market demand. The capacity indicates the maximum power the generator can produce. The load rate, calculated by dividing supply quantity by capacity, shows the operational status of power generation equipment. 

Suppose that $P_{h}^{D}$ is the day-ahead electricity price at the $h$ moment on $D$ day,   $C^{D}$ is the capacity on $D$ day, $Qs_{h}^{D}$ is the supply quantity at the $h$ moment on $D$ day, $Qd_{h}^{D}$ is the market demand quantity at the $h$ moment on $D$ day and $LR_{h}^{D}$ is the load rate at the $h$ moment on $D$ day. The forecast weather data for day $D+x$ include temperature and wind speed, denoted as $\hat{W}_{h}^{D+x} $, and the forecast market demand quantities published by third parties, denoted as $\hat{Qd}_{h}^{D+x} $. 

This study evaluates the economic revenue of day-ahead price forecasts for residents. Based on related work~\cite{zheng2020impact}, we formulate an optimization problem for the purchasing strategy, using given forecast prices and the energy storage systems. The cost savings from this strategy represent the economic revenue of price forecasts, and improved forecast accuracy enhances the revenue. The purchasing cost $C$ reaches its theoretical minimum $C_a$ if forecasts match actual prices perfectly. Without the strategy, the cost is $C_0$. The economic revenue $R$ equals $C_0 - C$. To assess the impact of forecast accuracy, we calculate $\alpha = \frac{R}{C_0 - C_a}$, reflecting the percentage of the maximum theoretical revenue achieved by the current forecast.

\section{Methodology}

\subsection{Framework Overview}

We notice that the day-ahead electricity price is determined by the supply and demand, so we directly use this correlation to forecast the price. Forecasting precise supply and demand curves is impractical due to the need to forecast individual trading decisions at various price levels. Obtaining such detailed data is challenging because of the private nature of these decisions. 

To address this issue, we introduce the assumption of the short-term stability of the supply curve. Using recent historical data, we reconstruct the supply-price correlation and obtain a forecast for the future supply curve. We fit the supply-price relationship using a simple \textbf{Co}rrelation-based \textbf{Pi}ecewise \textbf{Linear} Model (\textbf{CoPiLinear}). With this curve and supply volume forecasted by TSF models, we can forecast the price, as Figure~\ref{fig: overview} shows.

\subsection{Short-term Stability Assumptions}

\begin{figure*}[h]
\centering
\begin{minipage}[t]{0.3\textwidth}
\includegraphics[width=\textwidth]{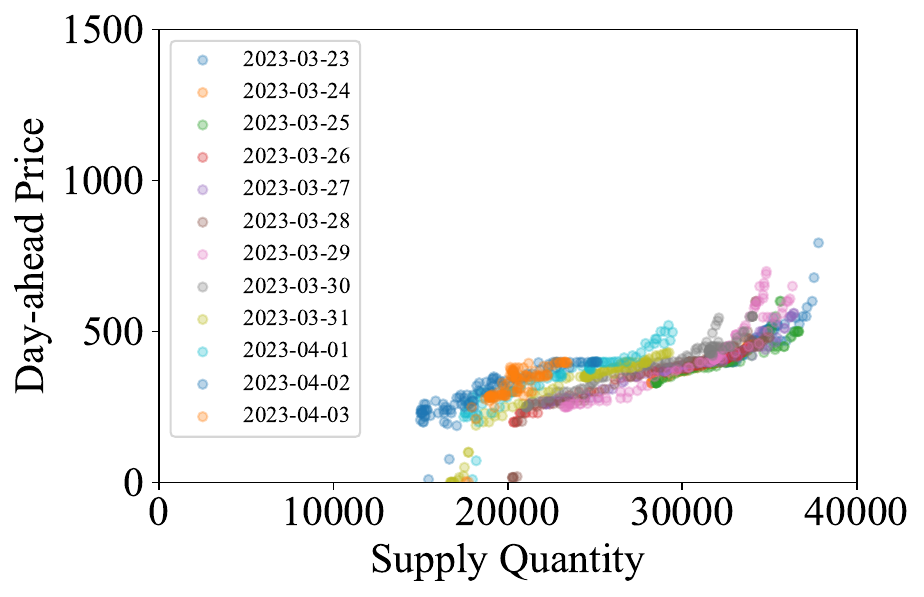}
\caption*{The supply curves, plotted with supply quantity on the x-axis, retain their shape over successive days, needing only slight left or right adjustments for alignment.}

\end{minipage}
\hspace{0.1cm}
\begin{minipage}[t]{0.3\textwidth}
\includegraphics[width=\textwidth]{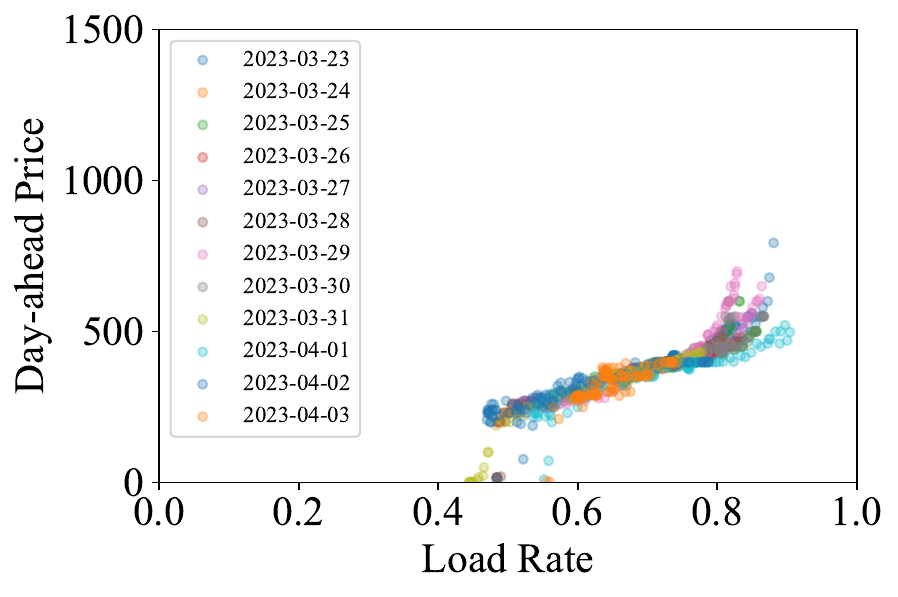}
\caption*{After normalizing supply by capacity, the aligned supply curves demonstrate short-term stability, with the x-axis representing the load rate.}

\end{minipage}
\hspace{0.1cm}
\begin{minipage}[t]{0.3\textwidth}
\includegraphics[width=\textwidth]{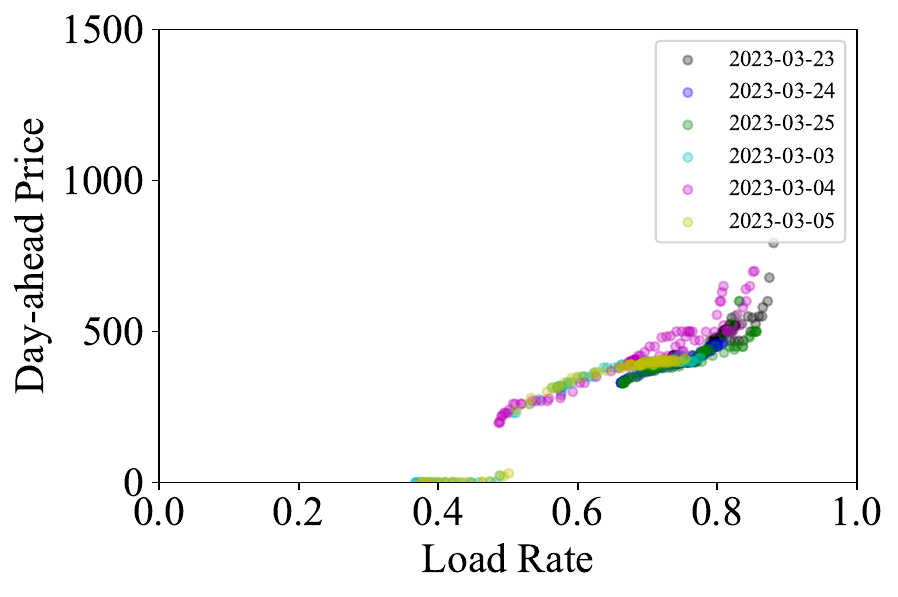}
\caption*{The disparity in supply curves increases when comparing dates separated by longer intervals.}

\end{minipage}
\caption{The supply curves of the Shanxi market show short-term stability within a period of adjacent dates.}
\label{fig:Temporal Invariance}
\end{figure*}

It is reasonable to assume the supply curve remains constant over a specific period. Commodities from different electricity power generators are homogenized, making the supply curve equivalent to the marginal cost~\cite{roberts1987large}. Notably, within a supplier, the marginal cost is primarily dictated by the load rate rather than the supply quantity. The load rate reflects the generator’s operating conditions, which directly impact costs~\cite{Roozbehani2022}: costs rise rapidly when transitioning from standby to startup, slow down upon reaching a stable segment, and surge again if a backup unit is activated. Due to managerial inertia, power plants typically do not alter operational units over several consecutive days, leading to a stable correlation between marginal cost and load rate during this period. More details about the short-term stability are listed in the Appendix B. 

Using the load rate instead of supply quantity as the horizontal axis of the supply curve, the correlation between price and load factor exhibits short-term stability over recent days. This conversion is straightforward by dividing the supply quantity by the capacity on that day.

As shown in Figure~\ref{fig:Temporal Invariance}, we selected a period from March 23, 2023, to April 3, 2023, within the Shanxi day-ahead electricity market to illustrate the short-term stability. The supply curve shape remains relatively stable across several consecutive days, requiring only minor shifts for approximate overlap. By using the load rate instead of supply quantity, we can align the supply curves of adjacent dates. This alignment reveals the short-term stability of the supply curve over consecutive days. Additionally, the similarity of supply curves decreases with longer intervals between dates.

\subsection{Fitting Supply Curve by CoPiLinear}

\begin{figure}[htbp]
  \centering
  \includegraphics[width=0.9\linewidth]{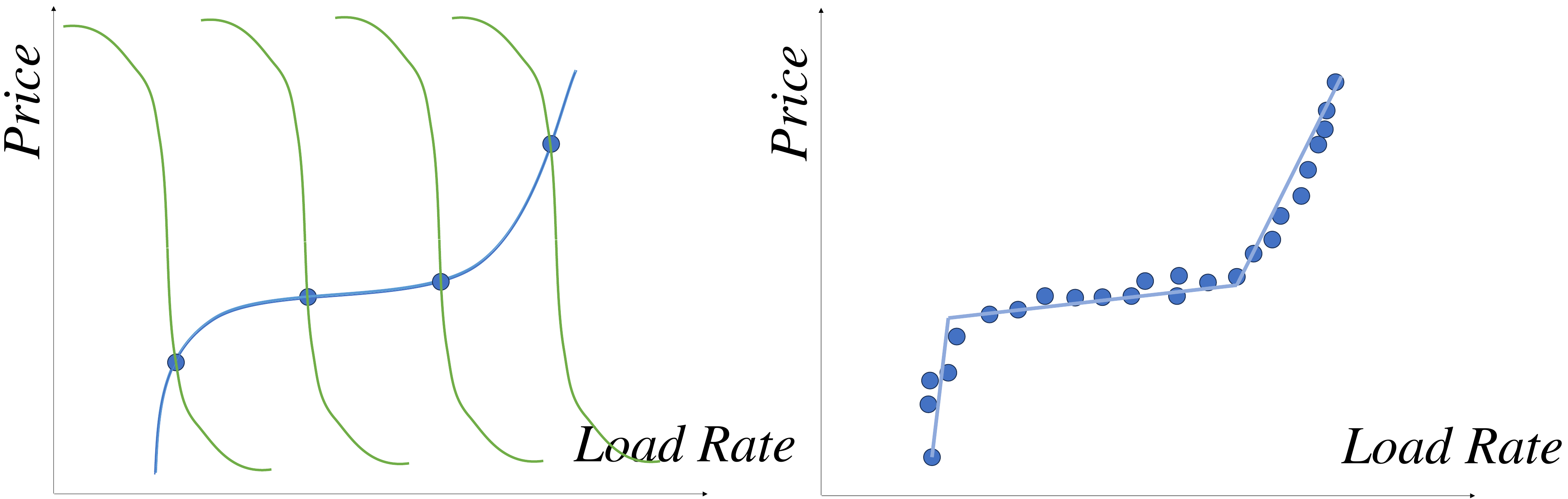}
  \caption{Left Image: The green curve shows varying demand over time, while the blue curve represents a stable supply. Their intersection points indicate market prices.
  Right Image: The scatter points represent supply-demand intersections over time, from which a supply curve can be derived.}
  \label{fig: intersection}
\end{figure}

Assuming short-term stability, we can forecast the supply curve by fitting it to recent days’ data. Our challenge is to determine the recent supply curve without data on willingness to buy or sell at each price. While the demand curve position varies significantly at different time slots, the supply curve remains relatively stable. We can get historical equilibrium day-ahead electricity prices and power supply volume at various trading time slots, corresponding to the intersections of the supply and demand curve. These intersections, all on the same supply curve, can be used to fit the supply curve in reverse, as shown in Figure~\ref{fig: intersection}.

We use the most relevant historical data to ensure accuracy. Supply curves are more similar between closer dates, as shown in Figure~\ref{fig:Temporal Invariance}. Therefore, our initial step is to identify and select the most suitable historical data. We model the supply curve for each day and examine these curves for deviations in shape. If a deviation exceeds a threshold, we retain only the data after the deviation.

The necessity for the most recent dates limits the amount of usable data, so it is imperative to employ a model that is relatively simple and easy to fit the supply curve. We represent the supply curve with a piecewise linear function, reflecting its phases with different slopes. Due to power plant generator characteristics~\cite{Roozbehani2022}, the supply curve consistently exhibited piecewise phases: rapid growth, steady growth, and then rapid growth again. Therefore, we employ the simplest form of a n-segment piecewise linear function to represent the supply curve, named CoPiLinear. The slopes and intercepts of the n lines are represented by \(w_1\), \(w_2\), \(\ldots\) , \(w_n\) and \(b_1\), \(b_2\), $\ldots$,  \(b_n\) respectively. And $LR_1^*$, $\ldots$,  $LR_{n-1}^*$ is the breakpoints of the piecewise linear function. 

We optimize the parameters using historical equilibrium points data. Throughout the day, we can observe prices and load rates at various transaction time slots. These correspond to scatter points on the supply and demand curve plane, note as \((LR_{1}, P_{1})\), \((LR_{2}, P_{2})\), \(\ldots\), \((LR_{N}, P_{N})\). The set of scatter points includes the closest $d$ days' data, each day has $k$ time slots. The goal is to fit these scatter points with CoPiLinear, minimizing the sum of distances from all points to the n-segment line. The final constraint equation ensures the continuity of the supply curve.

\begin{equation}
   \begin{aligned}
& \underset{n, w_1, w_2, \ldots, w_n, b_1, b_2, \ldots, b_n, LR_1^*, \ldots, LR_{n-1}^*}{\text{minimize}}
 \sum_{j=1}^N d_j^2 \\
& \text{subject to:}\\
& d_j = \left\{ \begin{array}{ll}
|P_j - (w_1LR_j + b_1)| & \text{if } LR_j \leq LR_1^* \\
|P_j - (w_2LR_j + b_2)| & \text{if } LR_1^* < LR_j \leq LR_2^*\\
\cdots &\cdots  \\
|P_j - (w_nLR_j + b_n)| & \text{if } LR_j > LR_{n-1}^*\\
j = 1, ..., N \\
\end{array} \right. \\
& w_{1} * LR_{1}^* + b_{1} = w_{2} * LR_{1}^* + b_{2}\\
& \cdots\\
& w_{n-1} * LR_{n-1}^* + b_{n-1} = w_{n} * LR_{n-1}^* + b_{n}
\end{aligned} 
\end{equation}



Fitting piecewise linear lines is a well-established problem with mature solutions. We use the Python package pwlf~\cite{jekel2019pwlf} to quickly approximate the supply curve. However, direct application that package on certain datasets may result in inaccuracies. To enhance fitting accuracy, we employ specific techniques detailed in Appendix C.

\subsection{Forecast related variables}

Electricity demand fluctuates significantly over time, making the demand curve challenging to forecast. By forecasting the supply curve, we can determine the price by identifying the demand quantity at the intersection of the supply and demand curves, without needing to forecast the entire demand curve.

Next, we need to forecast key supply-demand variables. The forecasted capacity data is used to adjust the horizontal axis of the CoPiLinear model from load rate to supply volume, creating the actual supply curve. The forecasted demand volume is then input into the CoPiLinear model to obtain the forecasted price.

These variables exhibit strong periodicity, making TSF models particularly effective, as detailed in Appendix D. Given their importance for the market, numerous mature TSF methods exist. Many service providers and market operators offer forecast data to aid market participants’ decision-making, so we directly utilize publicly available forecasts from third parties. In markets lacking forecasts for certain variable, such as capacity, we employ a simple model, Random Forest, to generate the necessary forecasts.

\section{Evaluation}

\subsection{Datasets and Baselines}

In our experiments, we utilized datasets from two distinct regions’ day-ahead electricity markets: Shanxi and ISO New England, some information is shown in Table~\ref{table:datasets_comparison}. The details are listed in the Appendix E.

\begin{table}[H]
\centering
\resizebox{.95\columnwidth}{!}{
\begin{tabular}{lcc}
\hline
& \textbf{Shanxi} & \textbf{ISO New England} \\ 
\hline
\textbf{Open access?} & Private & Public \\
\textbf{Time span} & 2023/03/01 - 2024/3/31& 2022/10/01 - 2023/12/31\\
\textbf{Test} & 2023/04/01 - 2024/3/31& 2023/01/01 - 2023/12/31\\
\textbf{Time granularity}& 15min& 1hour\\
\hline
\end{tabular}%
}
\caption{Shanxi and ISO New England Datasets Description.}
\label{table:datasets_comparison}
\end{table}

We selected two types of baselines: models relying solely on temporal dependence and those capturing variable correlations. For temporal dependence, we chose domain-specific methods such as SARIMA~\cite{zhao2017improvingsarima} and VMD-LSTM~\cite{xiong2023vmdlstm}, and generic TSF methods like DLinear~\cite{Zeng2022dlinear} and PatchTST~\cite{nie2022patchtst}. For variable correlations, we selected domain-specific methods including Linear~\cite{uniejewski2016automatedlinear}, SVM~\cite{prahara2022improvedsvm}, XGBoost~\cite{manfre2023hybridxgboost}, DNN~\cite{lago2018forecastingdnn}, and Lasso-RF~\cite{ludwig2015lassorf}, along with generic TSF methods such as Koopa~\cite{liu2023koopa}, TSMixer~\cite{chen2023tsmixer}, Informer~\cite{zhou2021informer}, Autoformer~\cite{wu2021autoformer}, FEDformer~\cite{zhou2022fedformer}, Crossformer~\cite{zhang2023crossformer}, and iTransformer~\cite{liu2023itransformer}. The variables used in these generic TSF models are exactly the same as those in the CoPiLinear, including historical price, capacity, demand data and forecast demand, weather data. 

To simulate real market applications, our model uses a daily rolling training method, adding new data each day for retraining. All baselines also adopt this method, updating the training and validation sets daily. Deep learning methods with short training sets tend to perform poorly. To improve accuracy, we use the longest possible training set and adjust hyperparameters during testing. More baseline details are in Appendix F.

\subsection{Experiment Settings and Evaluation Metric}

Our experiment platform is a server with 12 CPU cores (AMD Ryzen 9 7900X), and 32 GB RAM. Our GPU is NVIDIA GeForce RTX 4060 Ti 16 GB.

Based on related work~\cite{lago2021survey}, we select MAE and sMAPE to evaluate forecast accuracy. Reference~\cite{zheng2020impact} guides our choice of $R$ and $\alpha$ to assess the revenue of the forecast.  Detailed descriptions of the evaluation metrics are provided in Appendix G.

\subsection{Results and Analysis}

\begin{table*}[htbp]
\resizebox{\textwidth}{!}{%
\begin{tabular}{lcccclccccl}
\hline
 & \multicolumn{4}{c}{\textbf{Shanxi}} & & \multicolumn{4}{c}{\textbf{ISO New England }} & \\ 
  & \textbf{MAE}(\$/MWh)& \textbf{sMAPE} & \textbf{Revenue}(\$/year)& \textbf{$\alpha$} & \textbf{$MAE_{CS}$}& \textbf{MAE}(\$/MWh)& \textbf{sMAPE} & \textbf{Revenue}(\$/year)& \textbf{$\alpha$} & \textbf{$MAE_{CS}$}\\
\hline
 \textbf{SARIMA} & 13.0 & 0.383 & 8.40 $\times 10^7$ & 0.436  & 16.1& 8.97 & 0.211 & 3.43 $\times 10^7$ & 0.447  & 13.0\\
 \textbf{VMD-LSTM}& 12.7 & 0.357 & 8.48 $\times 10^7$ & 0.440  & 16.7& 21.7 & 0.510 & 1.94 $\times 10^7$ & 0.227  & 24.6\\
 \textbf{DLinear} & 12.6 & 0.370 & 8.40 $\times 10^7$ & 0.443  & 15.1& 12.0 & 0.292 & 2.97 $\times 10^7$ & 0.368  & 17.9\\
 \textbf{PatchTST} & 13.0 & 0.388 & 8.32 $\times 10^7$ & 0.437  & 16.3& 9.75 & 0.225 & 3.27 $\times 10^7$ & 0.423  & 17.2\\
\hline
 \textbf{Linear} & 18.0 & 0.440 & 7.25 $\times 10^7$ & 0.367  & 17.4& 31.7 & 0.669 & 1.27 $\times 10^7$ & 0.163  & 39.7\\
 \textbf{SVM} & 12.2 & 0.349 & 8.59 $\times 10^7$ & 0.448  & 15.6& 14.6 & 0.339 & 2.66 $\times 10^7$ & 0.316  & 19.5\\
 \textbf{XGBoost} & 11.2 & 0.317 & 8.76 $\times 10^7$ & 0.466  & 12.4& 9.57 & 0.219 & 3.32 $\times 10^7$ & 0.430  & 14.5\\
 \textbf{DNN} & 14.4 & 0.409 & 7.98 $\times 10^7$ & 0.415  & 15.8& 15.8 & 0.327 & 2.40 $\times 10^7$ & 0.295  & 20.1\\
 \textbf{Lasso-RF}& 11.4 & 0.355 & 8.76 $\times 10^7$ & 0.462  & 12.7& 8.80 & 0.196 & 3.52 $\times 10^7$ & 0.454  & 11.2\\
 \textbf{Koopa} & 13.2 & 0.387 & 8.24 $\times 10^7$ & 0.433  & 16.6& 10.5 & 0.246 & 3.26 $\times 10^7$ & 0.406  & 16.2\\
 \textbf{TSMixer} & 12.4 & 0.384 & 8.51 $\times 10^7$ & 0.435  & 15.8& 12.8 & 0.306 & 2.86 $\times 10^7$ & 0.349  & 19.5\\
 \textbf{Informer} & 16.6 & 0.414 & 7.43 $\times 10^7$ & 0.384  & 18.7& 11.4 & 0.285 & 3.00 $\times 10^7$ & 0.381  & 17.1\\
 \textbf{Autoformer} & 16.1 & 0.431 & 7.62 $\times 10^7$ & 0.391  & 22.7& 15.8 & 0.416 & 2.44 $\times 10^7$ & 0.297  & 24.4\\
 \textbf{FEDformer} & 14.2 & 0.411 & 8.40 $\times 10^7$ & 0.419  & 18.1& 14.9 & 0.377 & 2.64 $\times 10^7$ & 0.307  & 26.2\\
 \textbf{Crossformer} & 11.5 & 0.361 & 8.36 $\times 10^7$ & 0.462  & 12.7& 10.2 & 0.238 & 3.14 $\times 10^7$ & 0.412  & 15.0\\
 \textbf{iTransformer} & 13.1 & 0.394 & 8.32 $\times 10^7$ & 0.436  & 15.7& 10.3 & 0.258 & 3.29 $\times 10^7$ & 0.408  & 17.7\\
\hline
 \textbf{CoPiLinear} & \textbf{8.61}& \textbf{0.196}& \textbf{9.49 $\times 10^7$}& \textbf{0.512} & \textbf{10.8}& \textbf{7.77} & \textbf{0.175} & \textbf{3.62 $\times 10^7$} & \textbf{0.491}  & \textbf{9.78}\\
\hline
 \textbf{Co-POLY}& 9.30& 0.203& 9.30 $\times 10^7$&  0.500& 12.0& 10.6& 0.222& 3.17 $\times 10^7$& 0.400& 15.2\\
 \textbf{Co-EXP}& 9.68& 0.227& 9.30 $\times 10^7$&  0.494& 11.5& 10.6& 0.213& 3.06 $\times 10^7$& 0.398& 15.7\\
 \textbf{Co-XGB}& 10.1& 0.245& 9.20 $\times 10^7$&  0.484& 13.1& 12.1& 0.241& 2.85 $\times 10^7$& 0.361& 15.8\\
 \textbf{Co-MLP}& 11.8& 0.284& 8.68 $\times 10^7$&  0.457& 13.1& 12.7& 0.315& 2.82 $\times 10^7$& 0.352& 16.4\\
\hline
\end{tabular}
}
\caption{Results on Shanxi and New England Datasets.}
\label{table:combined_model_comparison}
\end{table*}

\subsubsection{Main results}

In Table~\ref{table:combined_model_comparison}, we present the main results. For consistency, prices in the Shanxi dataset are converted to US dollars using the exchange rate of 7.16 RMB to 1 USD as of the writing date. For the Shanxi dataset, CoPiLinear achieves the lowest MAE and sMAPE, outperforming the second-best by 22.9\% in MAE and 38.0\% in sMAPE. This superior forecast accuracy also boosts electricity purchase revenue, with CoPiLinear generating the highest revenue and alpha. To ensure reproducibility, we also test on the ISO New England dataset, where CoPiLinear again shows the lowest forecast error and achieving the highest revenue. Note that the electricity volume consumed by residents in Shanxi is approximately 3 times that of ISO New England, resulting in higher revenue.
 
\subsubsection{Analysis of forecast accuracy continuity}

\begin{figure}[h]
  \centering
  \includegraphics[width=\linewidth]{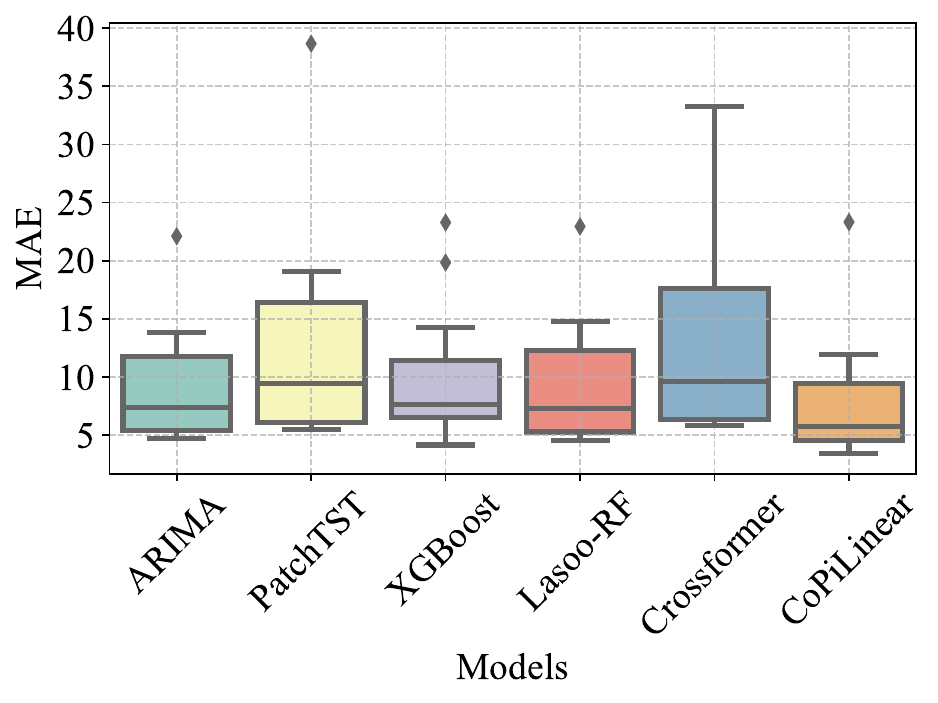}
  \caption{Shanxi market transaction process.}
  \label{fig: reliability_result}
\end{figure}

To ensure forecast reliability, we verify that the method maintains high accuracy across different times. We divide the ISO dataset’s test set (1 year) by month, select the top five methods with the best overall performance, and calculate each method’s monthly forecast error (MAE). A robust method should exhibit consistent accuracy each month. We plot the monthly forecast errors as box plots in Figure~\ref{fig: reliability_result}, revealing that CoPiLinear is the most stable. Additionally, higher MAEs are observed in summer months, coinciding with increased market energy consumption and price. 

\subsubsection{Ablation study 1}

For fitting the supply curve, we use a piecewise linear function, accurately capturing the different phases characterized by varying slopes. Alternative fitting methods include cubic functions, exponential functions, or models like MLP and XGBoost to implicitly learn the price-load correlation. To validate our approach, we tested several CoPiLinear variants: 1) Co-PLOY: cubic function; 2) Co-EXP: exponential function; 3) Co-XGB: XGBoost model; 4) Co-MLP: MLP model. These models differ only in the supply curve fitting function. Our ablation study results Table~\ref{table:combined_model_comparison} show that the original CoPiLinear model outperforms all variants on both datasets, confirming the effectiveness of using a piecewise linear function.

\subsubsection{Ablation study 2}

\begin{table}[htbp]
\resizebox{0.95\columnwidth}{!}{%
\begin{tabular}{lccccc}
\hline
   & \textbf{SARIMA}& \textbf{TimesNet}& \multicolumn{1}{c}{\textbf{XGBoost}} & \textbf{Random Forest}& \multicolumn{1}{c}{\textbf{Official Forecast}}\\ 
\hline
\textbf{RMSE} & 2030.4& 1532.2& 889.07&\textbf{839.34}& 1899.0 \\
\textbf{MAE} & 1450.8& 1182.3& 698.44&\textbf{651.41}& 1713.8 \\
\textbf{sMAPE} & 0.070585& 0.57557& 0.035612&\textbf{0.33041}& 0.091486 \\
\hline
\end{tabular}
}
\caption{Capacity Forecast Results.}
\label{table:capa_comparison}
\end{table}

When shifting the supply curve’s horizontal axis from quantity to load rate, we use a Random Forest model to forecast capacity. Given the limited capacity data (one entry per day), we select models with smaller training set requirements to compare forecasting accuracy: SARIMA, TimesNet and XGBoost. What's more, the ISO New England dataset includes the organization’s official capacity forecasts, allowing direct comparison. As shown in Table~\ref{table:capa_comparison}, our model achieves exceptional accuracy with only a 3\% sMAPE error, significantly outperforming the official forecast. Our Random Forest model’s higher accuracy compared to other models further validates the design of this module. This demonstrates the validity and effectiveness of our capacity forecasting method. 

\subsubsection{Case Study}

\begin{figure}[h]
  \centering
  \includegraphics[width=\linewidth]{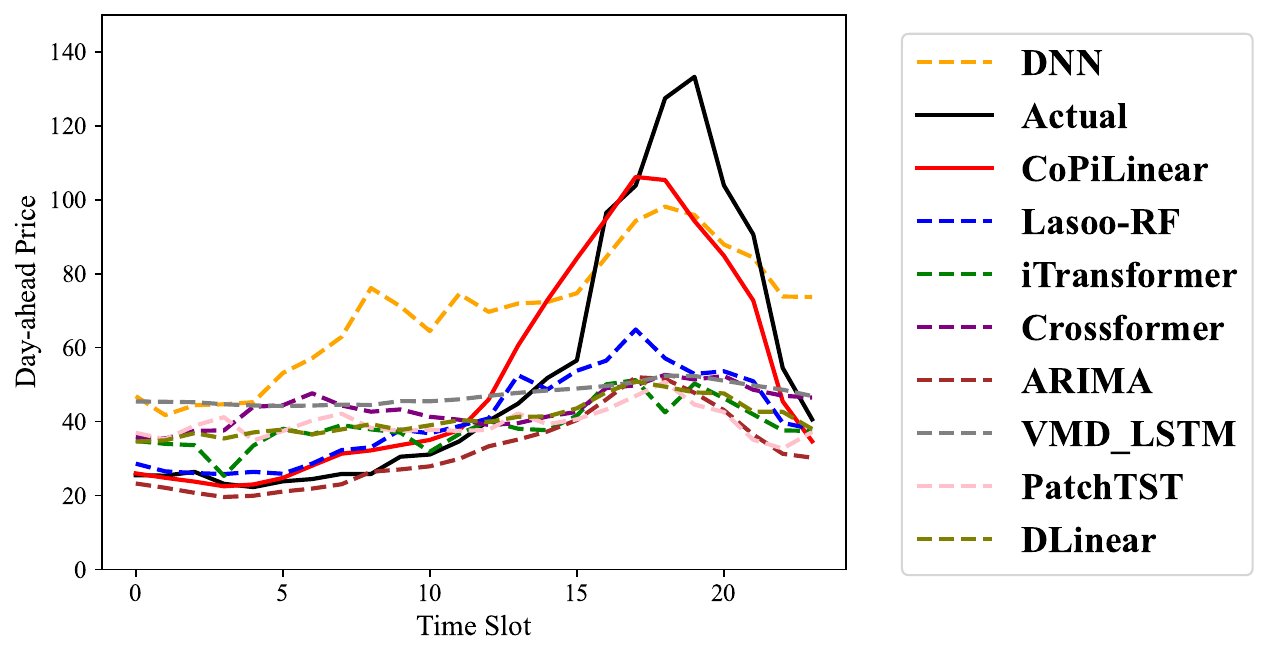}
  \caption{Shanxi market transaction process.}
  \label{fig: cs_result}
\end{figure}

Our method outperforms existing approaches in capturing the correlation between supply-demand and price. This allows for more sensitive adjustments to price forecasts based on forecasted demand or other variables, especially during significant supply and demand fluctuations. Figure~\ref{fig:shortcoming} analyzes the challenging high prices in the ISO New England market on July 17, 2023, due to a sharp increase in forecasted demand. Figure~\ref{fig: cs_result} shows the CoPiLinear forecast results align more closely with actual prices compared to other methods. The CoPiLinear forecasted price peaks a few phases earlier than the actual prices, it can be explained that it synchronizes completely with the demand surge, whereas the energy storage system in reality delays the impact of the demand surge on prices.

We validate this in two datasets. If the forecasted demand during the day exceeds the previous week’s maximum or falls below the minimum, we anticipate significant price fluctuations. We selected these dates and compared the MAE of various forecasting methods, as shown in the $MAE_{CS}$ column of Table~\ref{table:combined_model_comparison}. All methods showed increased MAE on these dates, with CoPiLinear achieving the highest accuracy in both datasets. 

\section{CONCLUSION}

This paper has highlighted the limitations of current day-ahead electricity price forecasting methods and introduced a novel approach to fit supply curve. By assuming short-term stability, CoPiLinear has significantly improved accuracy, benefiting residents. Rigorous testing on two electricity datasets shows that CoPiLinear outperforms top-tier methods, underscoring its potential to reduce residents' costs and enhance forecast reliability in day-ahead market operations. Future work will refine the model and explore its applicability to other markets. This achievement offers new perspectives for electricity price forecasting research and showcases the potential of artificial intelligence in the socio-economic field.

\bigskip


\bibliography{aaai25}

\section{Reproducibility Checklist}
 
This paper:

\begin{itemize}
    \item Includes a conceptual outline and/or pseudocode description of AI methods introduced (yes)
    \item Clearly delineates statements that are opinions, hypothesis, and speculation from objective facts and results (yes)
    \item Provides well marked pedagogical references for less-familiare readers to gain background necessary to replicate the paper (yes)
\end{itemize}
Does this paper make theoretical contributions? (yes)

If yes, please complete the list below.

\begin{itemize}
    \item All assumptions and restrictions are stated clearly and formally. (yes)
    \item All novel claims are stated formally (e.g., in theorem statements). (yes)
    \item Proofs of all novel claims are included. (yes)
    \item Proof sketches or intuitions are given for complex and/or novel results. (yes)
    \item Appropriate citations to theoretical tools used are given. (yes)
    \item All theoretical claims are demonstrated empirically to hold. (yes)
    \item All experimental code used to eliminate or disprove claims is included. (yes)
\end{itemize}
Does this paper rely on one or more datasets? (yes)

If yes, please complete the list below.

\begin{itemize}
    \item A motivation is given for why the experiments are conducted on the selected datasets (yes)
    \item All novel datasets introduced in this paper are included in a data appendix. (yes)
    \item All novel datasets introduced in this paper will be made publicly available upon publication of the paper with a license that allows free usage for research purposes. (partial)
    \item All datasets drawn from the existing literature (potentially including authors’ own previously published work) are accompanied by appropriate citations. (yes)
    \item All datasets drawn from the existing literature (potentially including authors’ own previously published work) are publicly available. (yes)
    \item All datasets that are not publicly available are described in detail, with explanation why publicly available alternatives are not scientifically satisficing. (yes)
\end{itemize}
Does this paper include computational experiments? (yes)

If yes, please complete the list below.

\begin{itemize}
    \item Any code required for pre-processing data is included in the appendix. (yes).
    \item All source code required for conducting and analyzing the experiments is included in a code appendix. (yes)
    \item All source code required for conducting and analyzing the experiments will be made publicly available upon publication of the paper with a license that allows free usage for research purposes. (yes)
    \item All source code implementing new methods have comments detailing the implementation, with references to the paper where each step comes from (yes)
    \item If an algorithm depends on randomness, then the method used for setting seeds is described in a way sufficient to allow replication of results. (yes)
    \item This paper specifies the computing infrastructure used for running experiments (hardware and software), including GPU/CPU models; amount of memory; operating system; names and versions of relevant software libraries and frameworks. (yes)
    \item This paper formally describes evaluation metrics used and explains the motivation for choosing these metrics. (yes)
    \item This paper states the number of algorithm runs used to compute each reported result. (yes)
    \item Analysis of experiments goes beyond single-dimensional summaries of performance (e.g., average; median) to include measures of variation, confidence, or other distributional information. (yes)
    \item The significance of any improvement or decrease in performance is judged using appropriate statistical tests (e.g., Wilcoxon signed-rank). (yes)
    \item This paper lists all final (hyper-)parameters used for each model/algorithm in the paper’s experiments. (yes)
    \item This paper states the number and range of values tried per (hyper-) parameter during development of the paper, along with the criterion used for selecting the final parameter setting. (yes)
\end{itemize}
 
\section{Appendix A}

High prices are mainly driven by high demand, which can help forecast these irregularities. However, TSF models struggle to capture the correlation between prices and supply-demand dynamics. According to the price formation mechanism, forecasted demand should significantly impacts forecasted price. Yet, even with substantial changes in demands, existing models show minimal price variation. This inconsistency with real market behavior indicates that these models do not adequately incorporate supply and demand variables, leading to poor forecasting performance. See Appendix A for detailed experiment results.

\begin{figure}[htbp]
\centering
\subfigure[Domain-specific models.]{
\includegraphics[width=0.22\textwidth]{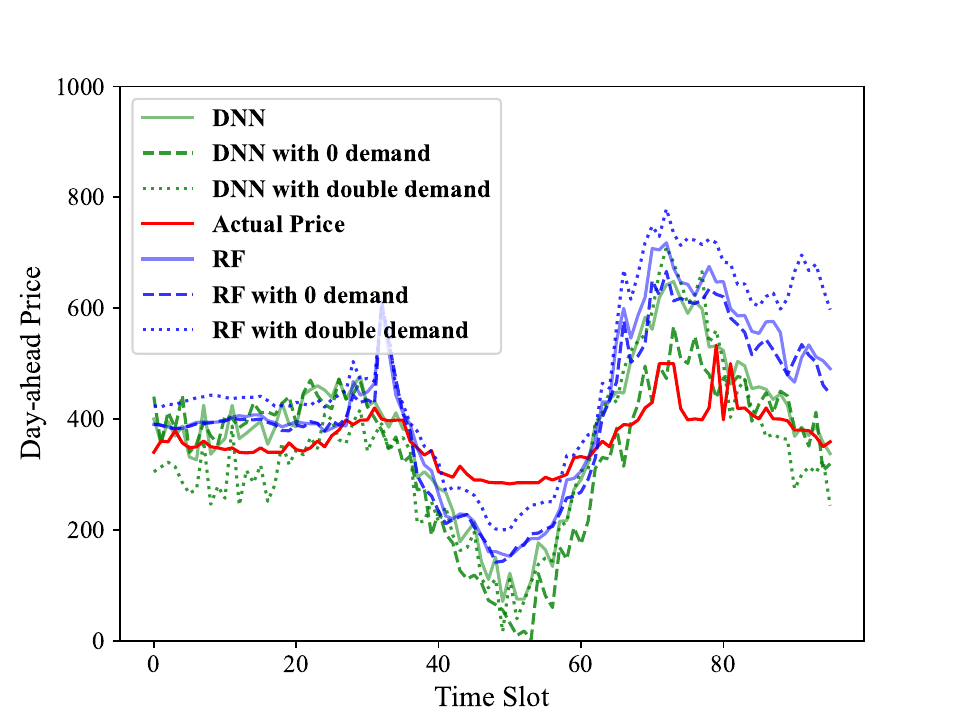}
\label{fig:shortcoming_a}
}
\subfigure[SOTA TSF models.]{
\includegraphics[width=0.22\textwidth]{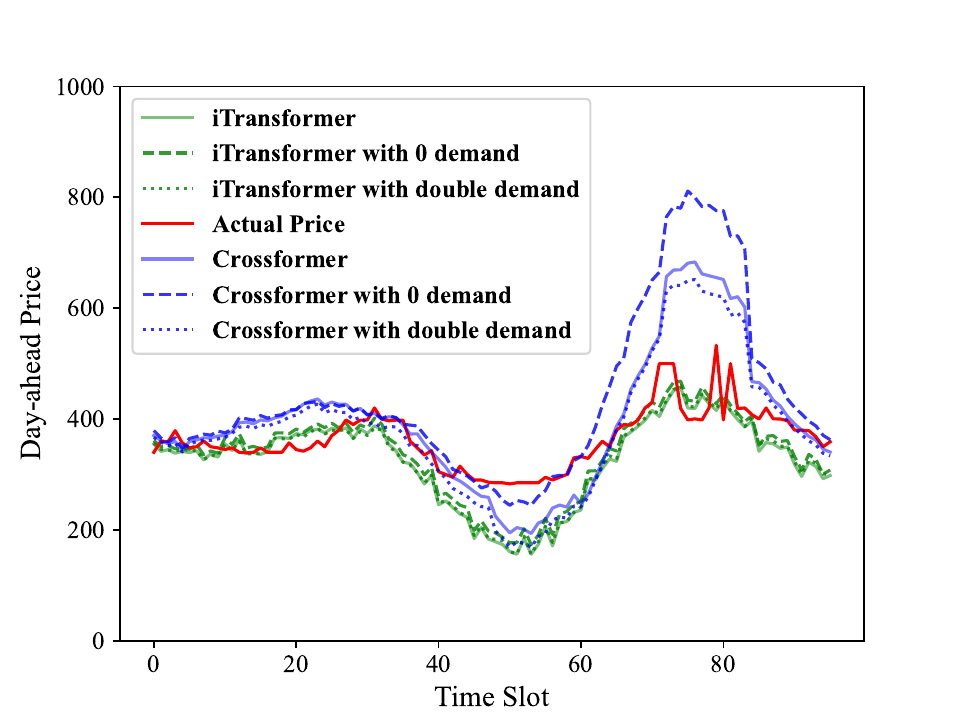}
\label{fig:shortcoming_b}
}
\caption{Forecasted prices with altered demand for January 15th, 2024.}
\label{fig:shortcoming1}
\end{figure}

To investigate this, we alter the demand for the target date, setting it to zero or doubling it, while keeping other inputs constant. Using data from Shanxi province for January 2024, we examine the impact of significant demand changes on the forecasted prices of four models: DNN, Lasso-RF, iTransformer, and Crossformer. The first two models are commonly used in day-ahead electricity price forecasting, while the latter two are state-of-the-art (SOTA) TSF models that consider relationships between variables. The sMAPE between the forecasted prices before and after the change is generally small, not exceeding 20\%, as shown in Table.~\ref{table:delta_smape}. Figure~\ref{fig:shortcoming1} shows the forecasted prices for January 15th, with minimal changes even when demand is significantly altered. Interestingly, forecasted prices often increase when demand is zero and decrease when demand is doubled, contradicting economic theory. 

\begin{table}[htbp]

\resizebox{\columnwidth}{!}{%
\begin{tabular}{lcccc}
\toprule
 & \multicolumn{1}{c}{\textbf{DNN}} & \multicolumn{1}{c}{\textbf{Lasso-RF}} & \multicolumn{1}{c}{\textbf{iTransformer}} &\textbf{Crossformer}\\
\midrule
\textbf{Zero Demand}& 0.20077 & 0.10006 & 0.055960 & 0.13900 \\
\textbf{Double Demand}& 0.27866 & 0.15165 & 0.073360 & 0.078306 \\
\bottomrule
\end{tabular}
}
\caption{sMAPE between the forecasted prices before and after demand change.}
\label{table:delta_smape}
\end{table}

\section{Appendix B}

The assumption of short-term stability is closely related to the modeling of thermal power generation. It is important to highlight that our methodology remains applicable in addressing the future’s needs, which include the anticipation of increased deregulation, the integration of renewable resources, and the implementation of energy storage solutions. Due to the high cost of thermal power generation, as long as it exists, it will serve as the marginal price of the supply curve. The instability of renewable energy generation and the inertia of social transition determine that thermal power generation is difficult to be completely eliminated in the foreseeable future, so the short-term stability assumption of the supply curve can still apply for a long time. As the proportion of renewable energy generation increases and energy storage technology develops, if thermal power generation has not been completely eliminated at this time, our method still applies according to the previous analysis, and even because of the progress of energy storage technology, the cost of supply during peak and trough periods will be more similar, which actually enhances our assumption of short-term stability (the supply curve is more like a horizontal line); if renewable energy occupies all power generation shares, the supply curve at that time may be more affected by storage scheduling costs, which also have short-term stability. Of course, these need to be studied in depth after the relevant technologies mature. It is worth mentioning that in the Shanxi market we analyzed, the proportion of renewable energy generation has already reached more than half, far higher than the global average level (30\%), and our method still works well. As for the increased market deregulation, our method comes from the principle of supply and demand and is still applicable in a free market. The ISO dataset in the experimental part is heavily deregulated markets, and the electricity price is determined by the transaction price. In summary, we believe that the assumption of short-term stability of the supply curve can be established in a wide range of time and space.

\section{Appendix C}

 \begin{itemize}
 \item \textbf{Data preprocessing.} There are instances when power generators undergo emergencies or maintenance, causing the supply curve to become highly irregular. This irregularity disrupts the correlation between price and load rate, making such dates challenging to forecast. Moreover, data from these days are unsuitable for future forecasting as they introduce noise. To address these issues, we preprocess the data before fitting the supply curve. We model the supply curve for each day using the relevant historical data. If the error between the modeled supply curve and the actual value exceeds a certain threshold, it indicates irregular bidding behavior on that day, and we discard the data for that day. This approach ensures that our model is based on the most relevant and accurate information. 
 \item \textbf{Weighting the data.} Power generators typically operate smoothly, which means that most of the historical data used for fitting falls within the second segment of the piecewise linear function we aim to fit (the stable part). The steeper first and third segments have fewer data points available for fitting. This scarcity makes it difficult for existing algorithms to accurately pinpoint the positions of the two breakpoints. To mitigate this, we increase the weights of the data points in the first and third segments during the fitting process. This adjustment enhances the effectiveness of the piecewise linear fitting. 
 \item \textbf{Anomaly Detection.} At times, nearly all historical data resides within the stable second segment of the piecewise linear function we aim to fit. Upon completing the piecewise linear fitting, we often observe that the results for the first and third segments of these dates, when fitted with a three-segment line, are quite extreme. Therefore, we conduct a post-fitting check. If we encounter clearly unreasonable data, such as a negative slope for the supply curve in the first and third segments, we substitute it with the slope of the second segment, which typically exhibits a more stable fitting outcome. 
\end{itemize}

\section{Appendix D}

\begin{figure}[htbp]
\centering
\subfigure[Load Rate]{
\includegraphics[width=0.2\textwidth]{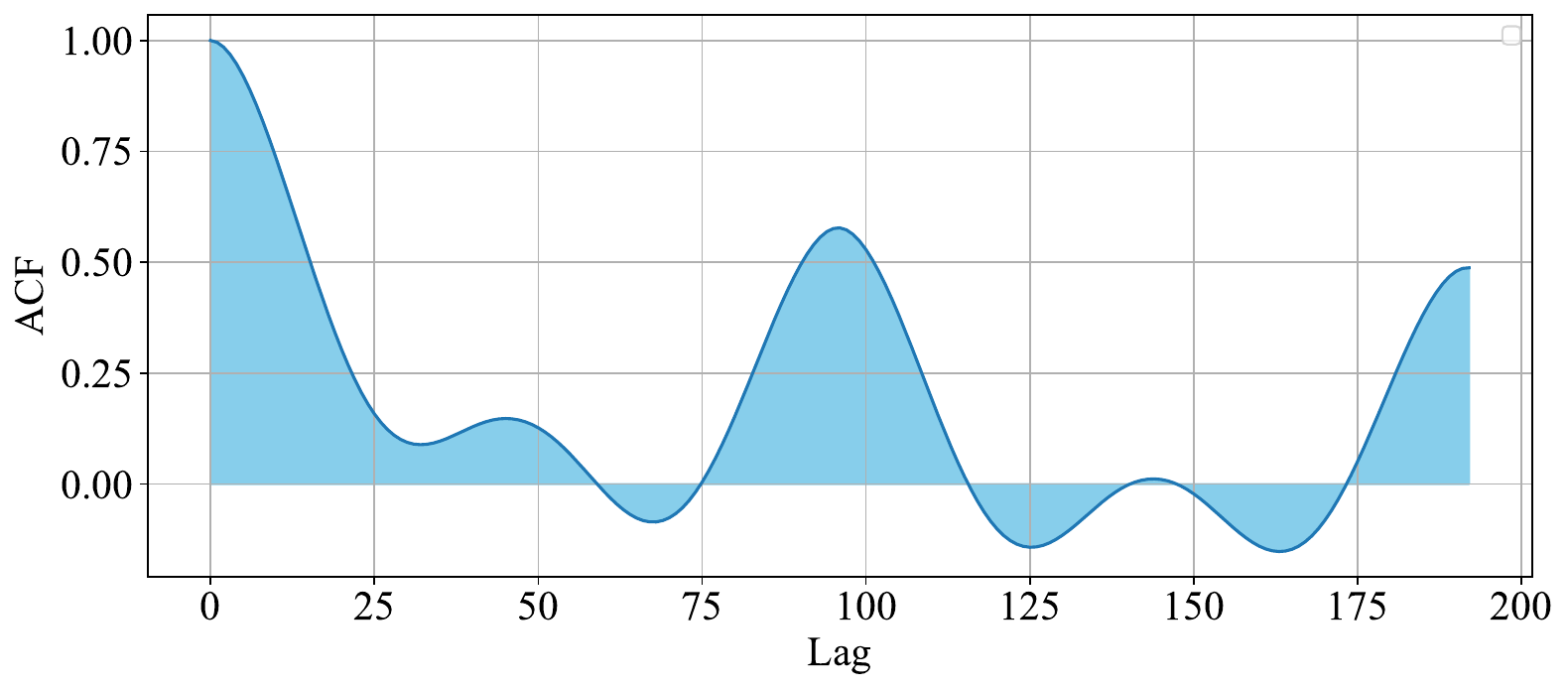}
\label{fig:ACF shanxi_a}
}
\hspace{0.1cm}
\subfigure[Day-ahead Price]{
\includegraphics[width=0.2\textwidth]{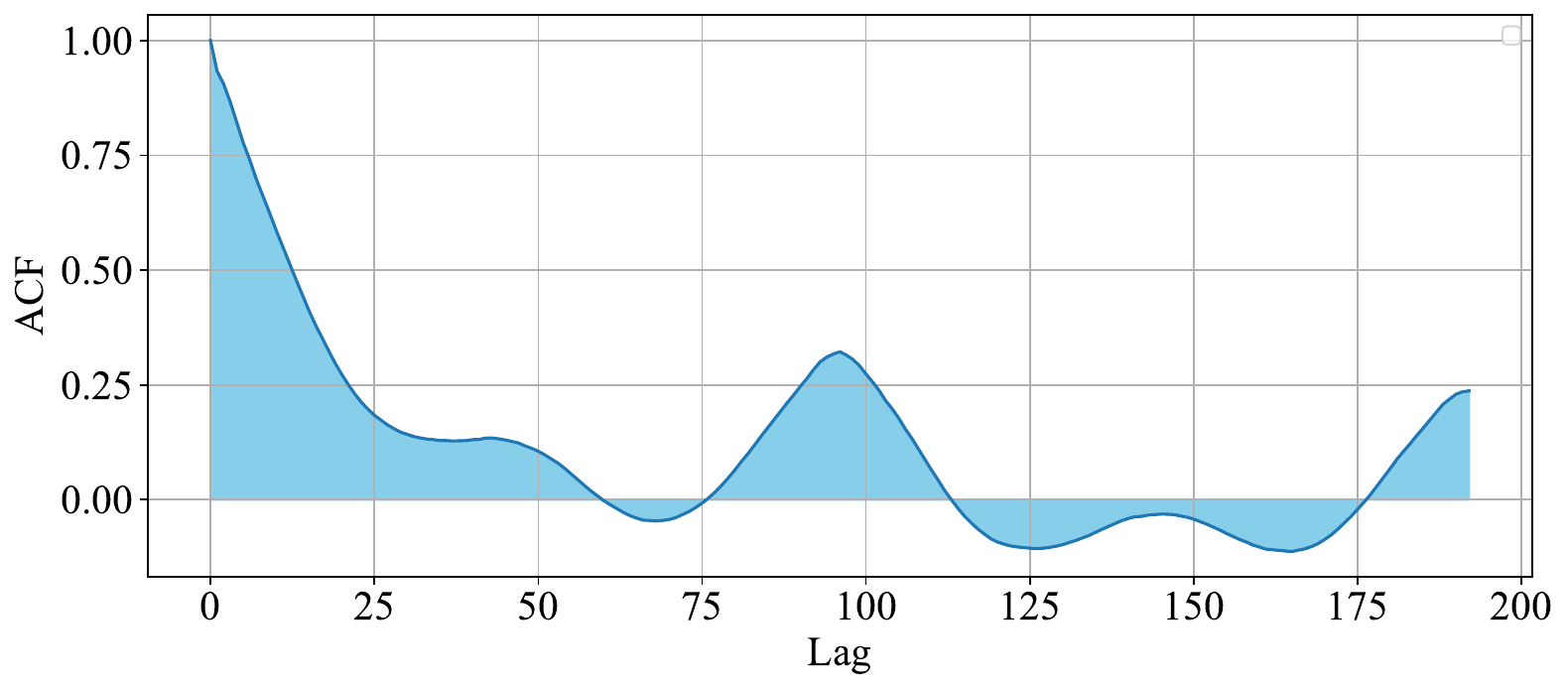}
\label{fig:ACF shanxi_b}
}
\caption{The ACF plot of the Load Rate and Day-ahead Price variable on Shanxi dataset, where every 96 lags on the x-axis represent one day.}
\label{fig:ACF shanxi}
\end{figure}

\begin{figure}[htbp]
\centering
\subfigure[Load Rate]{
\includegraphics[width=0.2\textwidth]{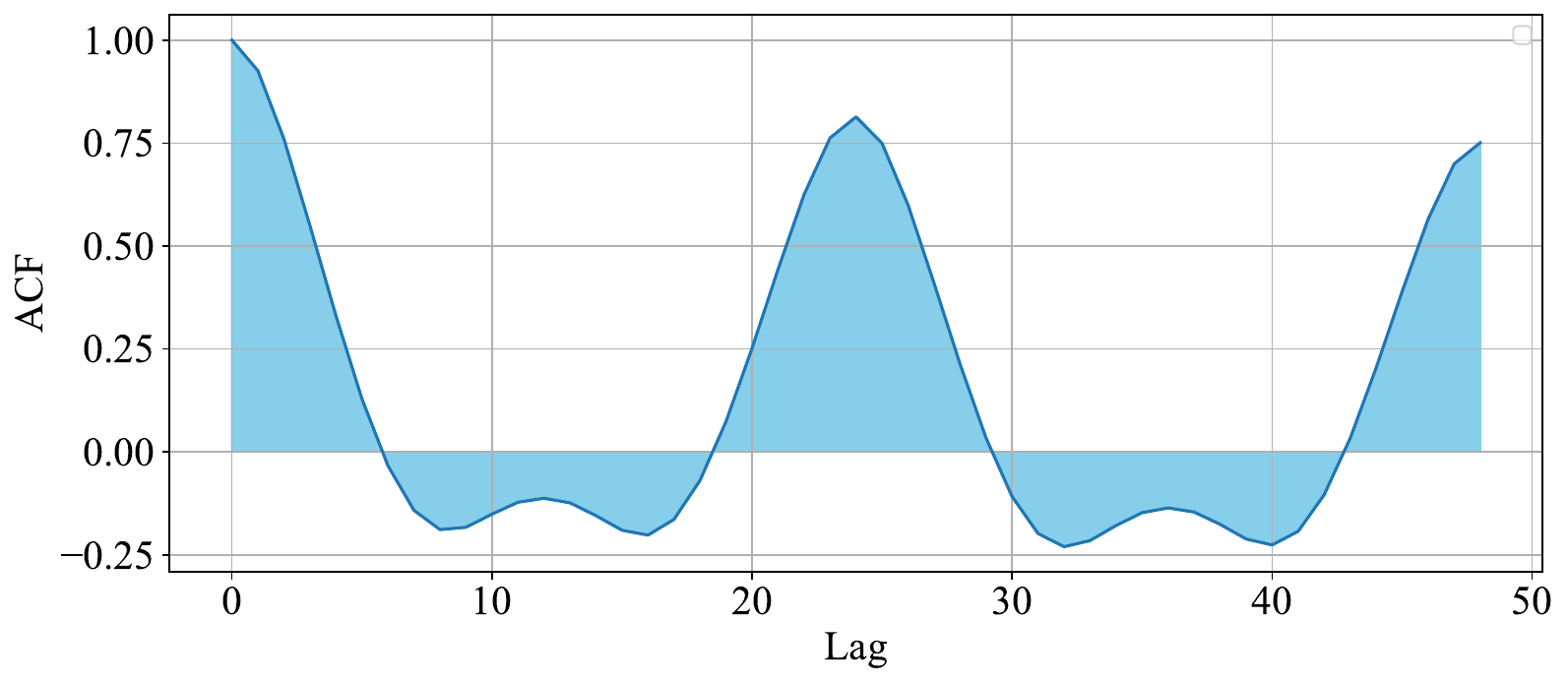}
\label{fig:ACF ISO_a}
}
\hspace{0.1cm}
\subfigure[Day-ahead Price]{
\includegraphics[width=0.2\textwidth]{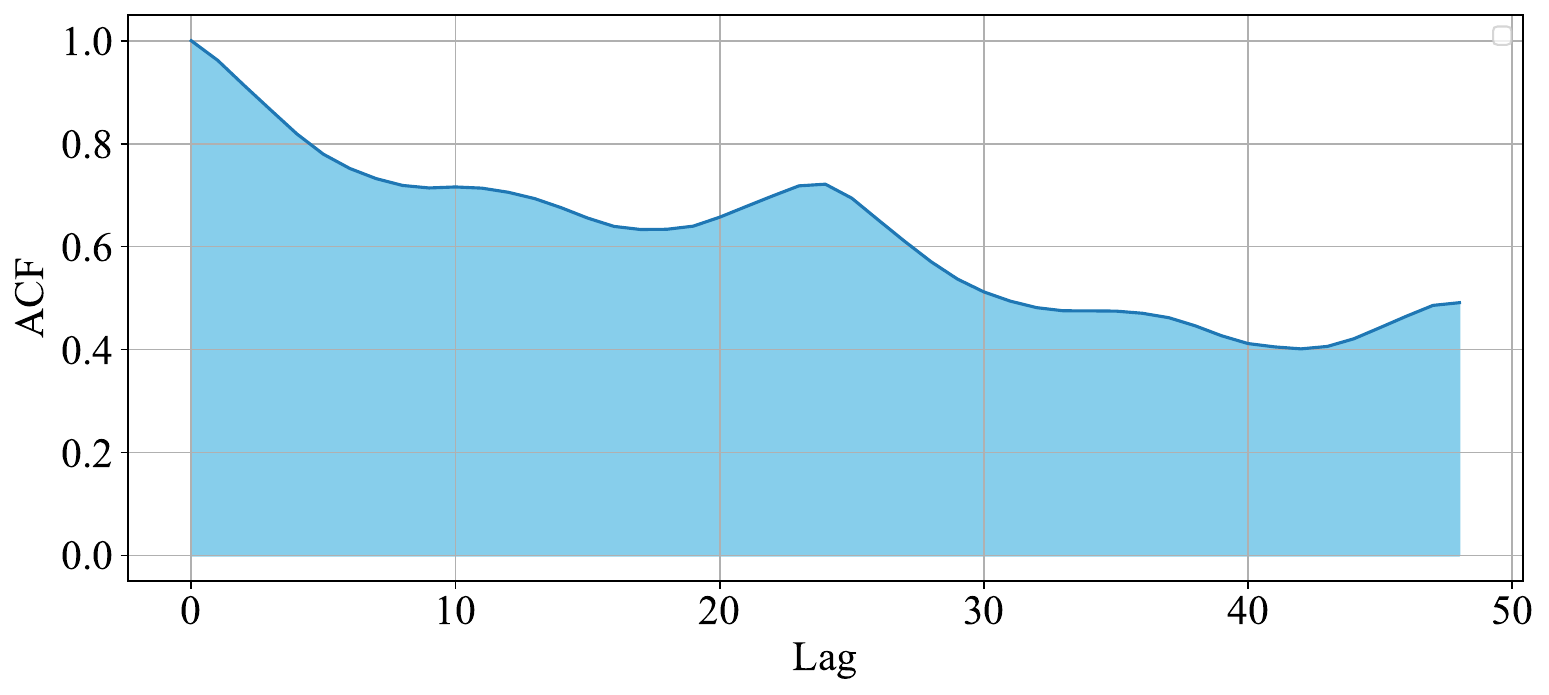}
\label{fig:ACF ISO_b}
}
\caption{The ACF plot of the Load Rate and Day-ahead Price variable on ISO New England dataset, where every 24 lags on the x-axis represent one day.}
\label{fig:ACF ISO}
\end{figure}

\begin{figure}[htbp]
\centering
\subfigure[Shanxi]{
\includegraphics[width=0.2\textwidth]{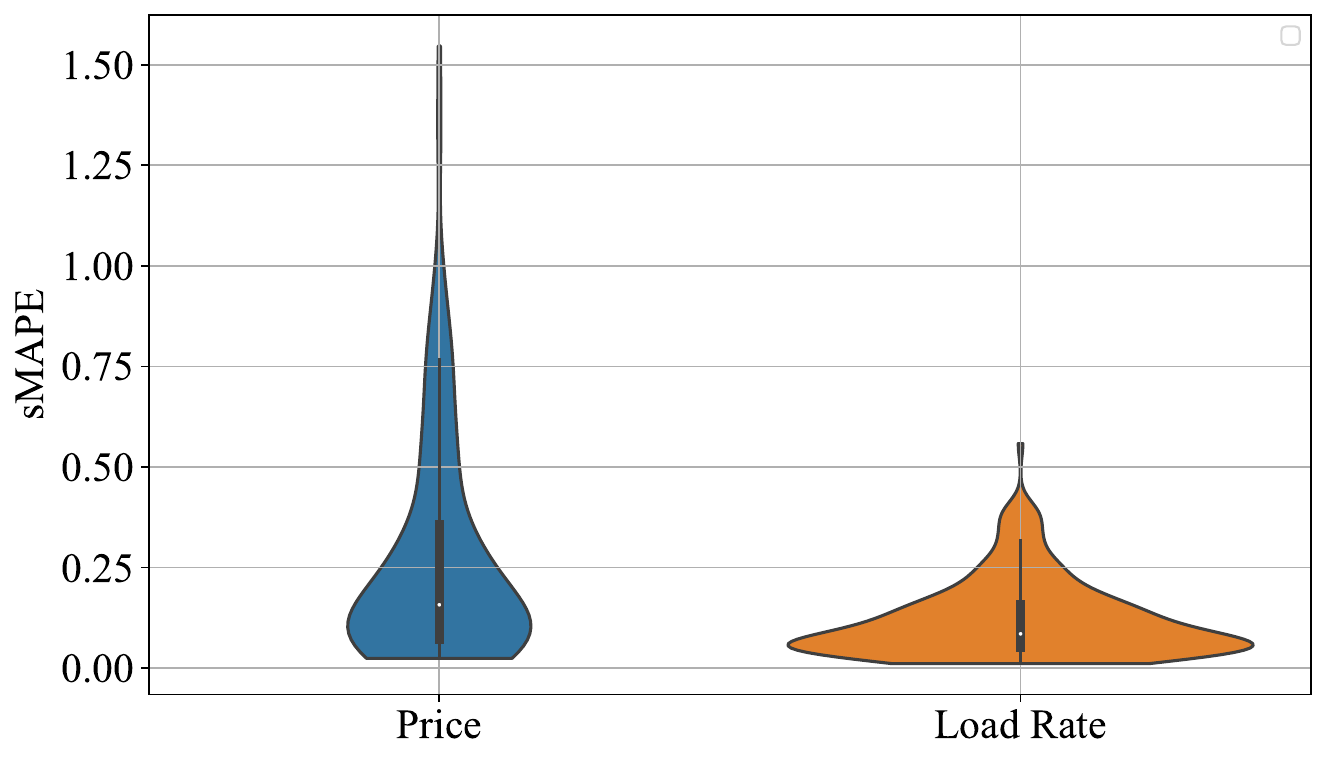}
\label{fig:smape_a}
}
\hspace{0.1cm}
\subfigure[ISO New England]{
\includegraphics[width=0.2\textwidth]{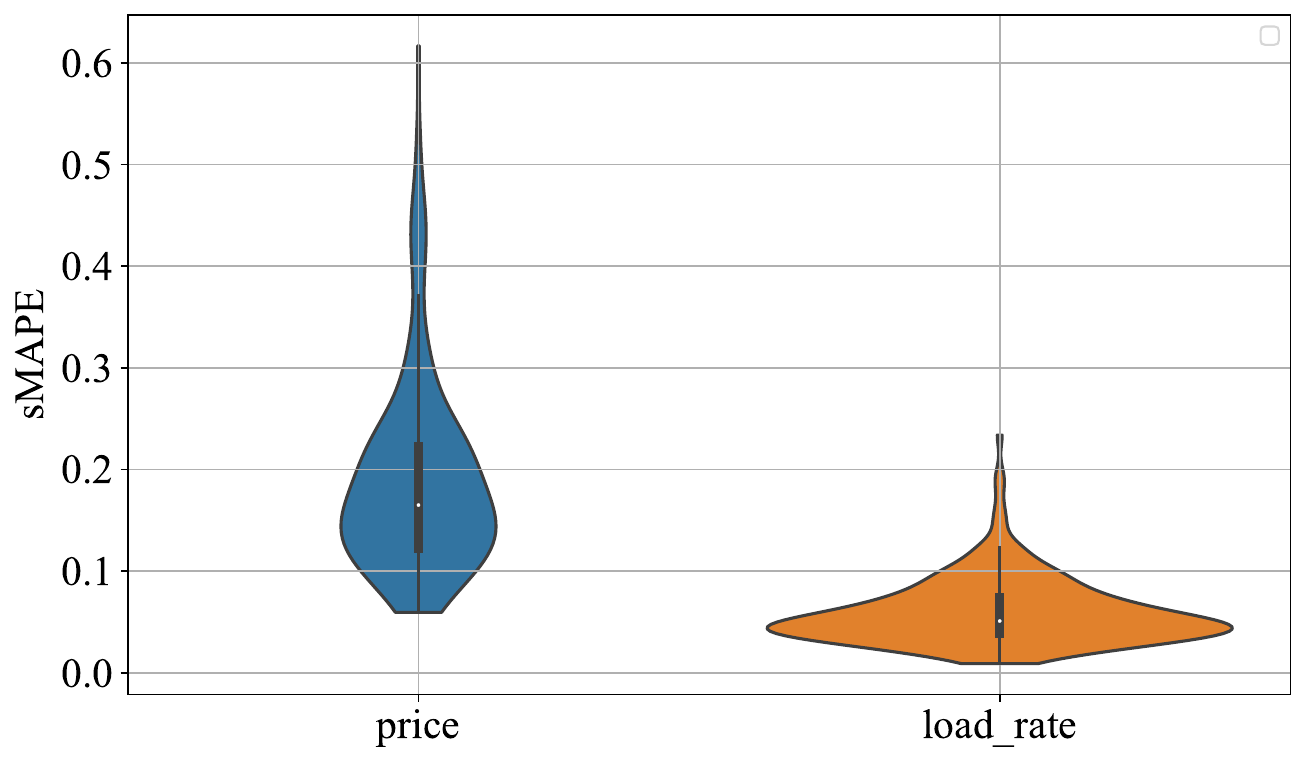}
\label{fig:smape_b}
}
\caption{Comparison of sMAPE for forecasting Load Rate and Day-ahead Price respectively on the Shanxi and ISO New England datasets using simple forecasting method.}
\label{fig:smape}
\end{figure}

We conducted a comparative analysis of the predictability of two variables, namely Load Rate and Day-ahead Prices, using Autocorrelation Function (ACF) and Symmetric Mean Absolute Percentage Error (sMAPE) derived from a simple forecasting method. The datasets used for this analysis were sourced from Shanxi and ISO New England. It’s important to note that both Load Rate and Day-ahead Prices exhibit daily periodicity. The ACF\_daily indicator, which measures predictability, shows that the larger the peak in the ACF curve, the stronger the predictability~\cite{Chen_2023liyue}. As depicted in Fig.~\ref{fig:ACF shanxi_a} and ~\ref{fig:ACF shanxi_b}, the ACF plot for Load Rate in the Shanxi dataset has a larger peak compared to that of the Day-ahead Price, indicating a higher ACF\_daily and thus, stronger predictability. This observation is consistent with the results from the ISO New England dataset, as shown in Fig.~\ref{fig:ACF ISO_a} and ~\ref{fig:ACF ISO_b}. We also employ a simple forecasting method, which involves forecasting the data for a target date using the data from the previous day. This method, effective for variables with daily periodicity, results in a sMAPE error. Fig.~\ref{fig:smape_a} and ~\ref{fig:smape_b} present the forecast results for Load Rate and Day-ahead electricity Prices using this method. In both datasets, the sMAPE for Load Rate forecasts was smaller. In conclusion, our analysis suggests that Load Rate is a more suitable variable for forecasting.

\section{Appendix E}

The dataset from Shanxi was collected from the official app, e-Trade, where the trading center publishes data. The time span of the Shanxi dataset ranges from March 1, 2023, to October 31, 2023. Starting from April 1st, we forecast the day-ahead electricity price for the target date once a day in a rolling manner. A trading time slot is established every 15 minutes, resulting in a total of 96 time slots throughout the day. The electricity price assigned to each time slot signifies the level of the day-ahead electricity price for that specific interval. In Shanxi day-ahead electricity market, the third party takes into account reports from electricity producers on their anticipated output and costs, forecasts overall electricity demand quantities, and considers constraints such as grid dispatch limitations and power plants' operational characteristics. Through intricate optimization calculations, the third party coordinates the power generation and dispatch plans of all parties, ensuring that societal electricity demand quantities are met by suppliers.  A unified day-ahead electricity price for the entire market is determined by optimization calculations based on market equilibrium price. Markets employing this method encompass the entire South Korean electricity market and several provincial electricity markets in China. In the day-ahead electricity price market of Shanxi, the trading center, acting as a third party, announces the day-ahead electricity price for D+1, as well as the regional weather forecast and demand quantity forecast from D+1 to D+5 at 7 p.m. on day D. The day-ahead market trading for D+2 is closed at 9:30 a.m. Based on these forecast values and their historical data, as well as historical data of the available generation capacity, we aim to forecast the electricity price for D+2 on day D during this time.

The dataset from ISO New England, which spans from October 1, 2022, to September 31, 2023, was procured directly from the official ISO website. Our objective was to forecast the day-ahead electricity price for the subsequent day, a task we performed daily in a rolling manner starting from January 1st. Trading time slots are systematically established at 1-hour intervals, culminating in a total of 24 slots over the course of a day. The electricity price allocated to each of these slots serves as an indicator of the anticipated electricity price for that specific duration. ISO New England’s day-ahead electricity market serves as a free and open marketplace for electricity producers and consumers, acting merely as a platform for transactions which can be concluded at any price. All transactions are recorded by a third party, providing a representative metric for the entire regional market. This approach is also adopted by markets such as the PJM, Nord Pool, and EPEX. At 8 a.m. on day D, the third party announces the day-ahead electricity price for that day, along with the regional weather forecast and demand quantity forecast from D+1 to D+3. The day-ahead market trading for D+1 is then closed at noon. Our goal was to forecast the electricity price for D+1 during this period of day D, based on these forecast values, their historical data, and the historical data of the available generation capacity.

In these two datasets, we use the supply and demand quantities data of thermal power generation to construct the supply and demand curve. The day-ahead electricity market comprises several key supplier categories: thermal, wind, solar, nuclear, and hydro power generation~\cite{bichler2022electricity}. The supply curves for these categories exhibit significant variations. 
The day-ahead electricity market can be approximated as a perfectly competitive market,~\cite{OpenStax2024,OpenStax2024_2,OpenStax2024_3,OpenStax2024_4} where highest marginal costs equals the price~\cite{hogan2022electricity}. In the electricity market, the highest marginal cost is typically associated with thermal power generation.~\cite{su2020quoting,hogan2022electricity} The ISO New England dataset corroborates this perspective. It documents the supply costs of various energy sources across 108,646 trading intervals from October 1, 2022, to November 11, 2023. The data reveals that in 80.1\% of these intervals, the marginal cost is tied to thermal power generation. This suggests a significant role of thermal power in influencing market dynamics. ~\cite{iso_marginal}This observation aligns well with conventional wisdom for several reasons:

 \begin{itemize}
 \item \textbf{Historical precedence.} Thermal power generation was established earlier than other forms of energy generation, providing a stable power supply~\cite{yang2022does,Farnoosh2022}.
 \item \textbf{Market dominance.} Thermal power holds a substantial share of the electricity market, reflecting the industry’s maturity~\cite{Liu2021}.
 \item \textbf{Resource allocation.} The cost of thermal power generation, which relies on higher-cost fossil fuels, exceeds that of photovoltaic, solar, and hydro power generation~\cite{Liu2020}.
\end{itemize}

Therefore, when examining the intersection of supply and demand curves in the day-ahead electricity market, it’s not necessary to consider all types of power generation manufacturers collectively. Instead, our attention should be concentrated on thermal power generation.

\section{Appendix F}

\begin{itemize}
 \item \textbf{TimesNet~\cite{wu2022timesnet}.} TimesNet, through its modular structure, decomposes complex time series changes into different periods. By transforming the original one-dimensional time series into a two-dimensional space and using CNN, it unifies the modeling of intra-cycle and inter-cycle changes.
 \item \textbf{Koopa~\cite{liu2023koopa}.} Koopa focuses on describing ubiquitous non-stationary time series. It models time series data from a dynamical perspective and naturally solves the problem of non-linear evolution in real-world time series through modern Koopman theory. 
 \item \textbf{Informer~\cite{zhou2021informer}.} Informer, an efficient transformer-based model for Long Sequence Time-series Forecasting (LSTF), addresses the Transformer’s issues of quadratic time complexity, high memory usage, and encoder-decoder architecture limitations. It features a ProbSparse self-attention mechanism for improved time complexity and memory usage, self-attention distilling for handling long input sequences, and a generative style decoder for fast long-sequence predictions.
 \item \textbf{Autoformer~\cite{wu2021autoformer}.} Autoformer, a novel architecture with an Auto-Correlation mechanism, improves long-term time series forecasting by efficiently discovering dependencies and aggregating representations, outperforming traditional Transformer models and achieving state-of-the-art accuracy across various applications. 
 \item \textbf{FEDformer~\cite{zhou2022fedformer}.} FEDformer is a novel method for time series forecasting that combines the Transformer model with seasonal-trend decomposition and frequency enhancement. This approach not only captures the global trend and detailed structures of time series, but also significantly improves prediction accuracy and efficiency. 
 \item \textbf{DLinear~\cite{Zeng2022dlinear}.} DLinear is a simple linear model. According to reports, its performance in the field of time series forecasting can be compared with Transformer-based models. 
 \item \textbf{iTransformer~\cite{liu2023itransformer}.} iTransformer is a novel Transformer-based architecture for time series forecasting. It considers different variables separately, with each variable being encoded into independent tokens. It uses attention mechanisms to model the correlation between different variables, and feed-forward networks to model the temporal correlation of variables, thereby obtaining a better sequence temporal representation. 
 \item \textbf{SARIMA~\cite{zhao2017improvingsarima,mchugh2019forecastingsarima}.}The SARIMA (Seasonal Autoregressive Integrated Moving Average) model is a statistical approach used for time series forecasting, which captures autocorrelation, differencing, and seasonality in the data. In day-ahead electricity price prediction, SARIMA can model the time-dependent structure and seasonality of the prices, providing accurate forecasts that are crucial for operational and strategic decisions in the energy market.
\item   \textbf{Linear~\cite{uniejewski2016automatedlinear,lago2021survey}.} Linear models are models that assume a linear relationship between the input and output variables. They are widely used in electricity price forecasting. 
  \item \textbf{SVM~\cite{che2010shortsvm, wang2017robustsvm, prahara2022improvedsvm}.} SVM is a powerful supervised learning algorithm that efficiently perform a non-linear classification using what is called the kernel trick, implicitly mapping their inputs into high-dimensional feature spaces. SVM models are also used in electricity price forecasting. 
  \item \textbf{XGBoost~\cite{manfre2023hybridxgboost,xie2022forecastingxgboost}.}The XGBoost (Extreme Gradient Boosting) model is a machine learning technique that uses gradient boosting framework for regression and classification problems. In day-ahead electricity price prediction, XGBoost can handle non-linear relationships between features and target variable, and it’s robust to outliers, making it a powerful tool for predicting prices with high accuracy.
 \item \textbf{DNN~\cite{yamin2004adaptivednn,lago2018forecastingdnn,darudi2015electricitydnn}.} The DNN is a multi-layer feed forward network that uses a multivariate framework. There are many DNN-based models in electricity price forecasting.
 \item \textbf{LASSO-RF~\cite{ludwig2015lassorf}.} This model utilizes LASSO for feature selection to enhance the accuracy of electricity price forecasting. A case study for electricity price forecasting is presented, comparing different models. Based on the evaluation of forecasting accuracy, the final model used for price forecasting is Random Forest, which can automatically select important variables and handle non-linear relationships. 
 \item \textbf{VMD-LSTM~\cite{xiong2023vmdlstm}.} This model includes three strategies: an Adaptive Copula-Based Feature Selection (ACBFS) algorithm for input feature selection, a new signal decomposition technique based on a decomposition denoising strategy, and a Long Short-Term Memory (LSTM) model for forecasting.
 
\end{itemize}
The detaied experiment hyperparameters can be seen in supplementary material and code.

\section{Appendix G}

In the field of electricity price forecasting, the most widely used metrics to measure the accuracy of point forecasts are the Mean Absolute Error (MAE), the Root Mean Square Error (RMSE), and the Mean Absolute Percentage Error (MAPE). In particular, in most electricity trade applications, the underlying risk, profits, and costs depend linearly on the price and on the forecasting errors. Hence, linear metrics represent better than quadratic metrics the underlying risks of forecasting errors~\cite{lago2021survey}.

However, MAPE values become very large with prices close to zero (regardless of the actual absolute errors), the MAPE is usually dominated by the periods of low prices and is also not very informative. While the Symmetric Mean Absolute Percentage Error (sMAPE), is a commonly used measure of accuracy of predictive models. Compared to MAPE, sMAPE has better symmetry and stability when actual values are close to zero.

The formula for sMAPE is:

\begin{equation}
\begin{aligned}
sMAPE = \frac{100\%}{n} \sum_{t=1}^{n} 2 \frac{|Y_t - \hat{Y}_t|}{|Y_t| + |\hat{Y}_t|}
\end{aligned}    
\end{equation}

where $Y_t$ is the actual value, $\hat{Y}_t$ is the predicted value, and $n$ is the number of observations.

\end{document}